\documentclass{article}

\usepackage{arxiv}

\usepackage[utf8]{inputenc} 
\usepackage[T1]{fontenc}    
\usepackage{hyperref}       
\usepackage{url}            
\usepackage{booktabs}       
\usepackage{amsfonts}       
\usepackage{nicefrac}       
\usepackage{microtype}      
\usepackage{lipsum}
\usepackage{graphicx}
\usepackage{orcidlink}
\graphicspath{ {./images/} }

\newcommand{\imgcell}[1]{\includegraphics[width=\imgwidth]{#1}}
\newcommand{\methodcell}[1]{\raisebox{0.38cm}[0pt][0pt]{\small #1}}
\usepackage{pgf}
\usepackage{tikz}
\usepackage[export]{adjustbox}
\usepackage{multirow}
\usepackage{amsmath}
\usepackage{lipsum}
\usepackage{caption}

\title{SpectralCTGaussians: Projection-Domain Reconstruction and Basis Material Decomposition for Spectral CT using 3D Gaussian Splatting}

\author{
Reinout Vos \\
Delft University of Technology, Delft, Netherlands \\
\texttt{R.H.A.Vos@student.tudelft.nl} \And
Saptarshi Neil Sinha \\
Fraunhofer IGD, Darmstadt, Germany \\
\texttt{saptarshineilsinha@gmail.com} \And
Michael Weinmann \\
Delft University of Technology, Delft, Netherlands \\
\texttt{m.weinmann@tudelft.nl}
}

\begin{document}
\maketitle
\renewcommand{\thefootnote}{\fnsymbol{footnote}}
\footnotetext[1]{Equal contribution}

\begin{abstract}
Spectral computed tomography (CT) extends conventional CT by measuring attenuation
across multiple energy channels, allowing improved modeling of physical X-ray
interactions and energy-dependent material behavior and leading to richer scene
understanding. We present a novel method for spectral CT reconstruction and basis
material decomposition using 3D Gaussian Splatting by adding per-Gaussian basis
material fractions to the set of learnable parameters, which together with a set of
energy-dependent basis functions define the attenuation across the full spectral
range. The basis functions represent various physical attenuation models such as photoelectric absorption and Compton scattering, and are jointly optimized across all energy channels through a differentiable polychromatic forward model, with material decomposition performed via mean-shift clustering of the resulting coefficients. We evaluate our method on a baseline real-world dataset as well as a synthetic dataset that we introduce, comparing against traditional reconstruction algorithms and state-of-the-art learning-based CT reconstruction methods. Our approach outperforms all traditional baselines in novel view synthesis and achieves the best PSNR among all compared methods for spectral CT volume reconstruction, while describing all energy channels with a single shared representation that requires a number of Gaussians comparable to single-channel Gaussian splatting-based CT reconstruction approaches. For basis material decomposition, no traditional or learning-based baseline offers one-step decomposition with direct RGB material segmentation, and our method additionally recovers the photoelectric basis with higher PSNR than traditional pipelines.
\end{abstract}

\section{Introduction}
\label{sec:introduction}
Computed Tomography (CT) is one of the most widely used imaging modalities in medicine and industrial inspection, reconstructing the three-dimensional structure of an object from X-ray attenuation measured by a rotating source-detector pair~\cite{NIBIB_CT, FDA_CT}. Conventional CT records only a single energy-integrated intensity per detector pixel, so materials with similar effective attenuation are difficult to separate~\cite{e08c3d69dffe4fb4a397914be42b3101}. Spectral CT resolves the detected photons into multiple energy channels and, hence, allowing more accurate energy-level-aware scene analysis~\cite{doi:10.1148/radiol.2015142631, So_spectralCTPrinciples, GREFFIER2023167}. Since photoelectric absorption and Compton
scattering exhibit different dependencies on photon energy, energy-resolved measurements constrain
the material composition of the object rather than only its effective attenuation.
This enables basis material decomposition, in which the attenuation field is expressed
as a combination of physically meaningful basis functions~\cite{alvarez1976_energySelectiveCT},
and it supports applications such as contrast agent quantification, virtual removal of
bone or iodine, and improved noise modeling~\cite{doi:10.1148/radiol.2016160890, Yu_2016}.
In some protocols the spectral information also reduces the need for repeated or
multiphase acquisitions, which can lower radiation or contrast agent
dose~\cite{fornaro2011_multienergyCT, Reimer1617, 10.1259/bjr.20170290}. This comes at a cost, since the inverse problem grows in dimensionality, as each channel receives fewer photons and is therefore noisier, and the channels are coupled through the polychromatic forward model~\cite{Bousse_2024}. Existing methods address this only partially. Analytical algorithms such as FDK~\cite{Feldkamp:84} treat each channel independently, model-based iterative methods
capture the spectral physics but are expensive and rely on hand-crafted regularization,
two-step pipelines propagate reconstruction errors into the decomposition, and
learning-based methods are fast but typically train one model per channel and generalize
poorly.

Recently, NeRF~\cite{mildenhall2020nerfrepresentingscenesneural} and 3D Gaussian Splatting (3DGS)~\cite{kerbl20233dgaussiansplattingrealtime} have been adapted to CT. Neural fields have been extended to spectral data~\cite{shi2024raydrivenspectralctreconstruction, xie2024acindsparsectreconstruction, 10794200} but remain slow due to dense ray sampling, whereas 3DGS-based methods~\cite{r2_gaussian, lin2024learning3dgaussiansextremely, cai2024radiativegaussiansplattingefficient} are considerably faster, provide an explicit differentiable volume and regularize sparse-view acquisitions well, yet assign only a single density per Gaussian and cannot model energy-dependent attenuation. Spectral variants of 3DGS exist for optical rendering~\cite{sinha2024spectralgaussianssemanticspectral3d, thirgood2024hypergshyperspectral3dgaussian}, but model view-dependent appearance rather than X-ray attenuation. To the best of our knowledge, no 3DGS-based approach~\cite{kerbl20233dgaussiansplattingrealtime} supports a spectral CT representation with one-step basis material decomposition. We introduce the first such method, and in light of these improvements, the main contributions of this paper are:
\begin{itemize}
    \item We introduce a \textbf{spectral Gaussian representation} for CT, which extends Gaussian-splatting-based CT reconstruction from a single density field to a spectral representation in which energy-dependent basis coefficient values are included directly in the optimization process.
     \item We extend the differentiable CT rasterizer and voxelizer to compute basis-coefficient line integrals and embed them in a polychromatic forward projection model, so that all energy channels are \textbf{optimized jointly in a single efficient pass, using a number of Gaussians comparable} to that of a single-channel representation.
    \item We enable \textbf{one-step basis material decomposition} by optimizing the basis
    coefficients directly from the spectral projections, following the physics-based
    photoelectric and Compton models~\cite{alvarez1976_energySelectiveCT}. We complement
    them with a \textbf{learnable K-edge basis} that captures contrast agents such as
    iodine and metals, and we segment the resulting coefficients with mean-shift
    clustering~\cite{comaniciu2002mean} into material maps with direct RGB output.
    \item Finally, we introduce a synthetic spectral CT dataset with ground-truth material information, enabling quantitative evaluation of the decomposition.
\end{itemize}

\begin{figure*}
    \centering
    \includegraphics[width=\linewidth]{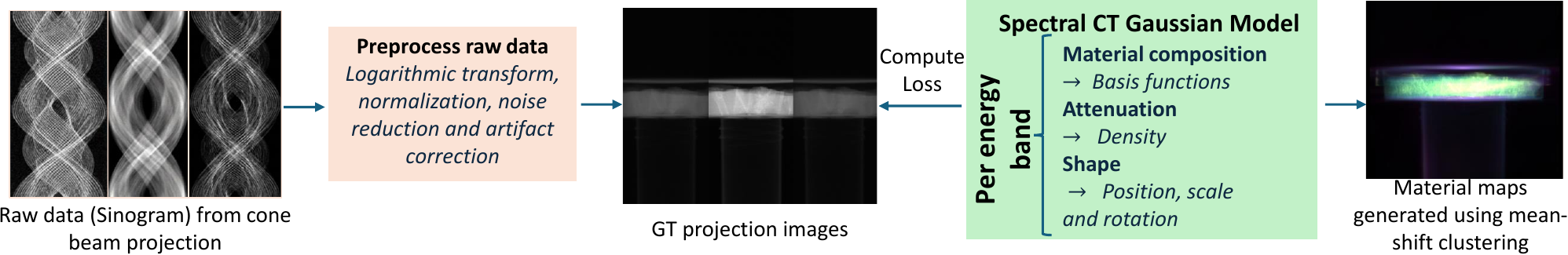}
    \caption{Overview of SpectralCTGaussians. Cone-beam projections optimize a 3D 
Gaussian Splatting model in which each Gaussian encodes geometry, density and 
basis material coefficients, supervised by the difference between rendered and 
measured log-projections across energy channels. Mean-shift clustering of the 
optimized coefficients yields material maps.}
    \label{fig:pipeline_spectral_ct_gaussians}
\end{figure*}
\section{Related Work}
\label{sec:related_work}
\noindent\textbf{Channel-wise reconstruction.}
The simplest way to handle spectral data is to reconstruct every energy channel independently as a conventional CT problem. Analytical algorithms such as FBP~\cite{osti_4493713} and its cone-beam extension FDK~\cite{Feldkamp:84}, as well as iterative solvers such as ART~\cite{GORDON1970471}, SART~\cite{ANDERSEN198481} and SIRT~\cite{GILBERT1972105}, are widely used because they are fast and easy to apply. However, they ignore the polychromatic nature of the beam and the correlation between channels, so beam hardening and cross-channel effects remain unmodeled, which makes them mainly useful as a quick initialization.
\noindent\textbf{Model-based iterative reconstruction.}
Model-based iterative reconstruction (MBIR) instead formulates spectral CT as a physics-based optimization problem combining a forward model, a statistical data fidelity term and a regularizer~\cite{Long2014}. Early work focused on priors that make sparse and low-dose acquisitions tractable, using low-rank and sparsity models~\cite{gao2011multi}, dictionary learning~\cite{zhao2012dual, zhao2013tight}, local low-rank patches~\cite{6985637} and multi-channel weighted least squares~\cite{Sawatzky2014ProximalADMM}. Tensor-based methods treat the data as a space-space-energy tensor to decouple spatial and spectral priors~\cite{6737273, zhang2017_tensorDLspectralCT, 7565539}, and dictionary learning has also been applied to basis material coefficients~\cite{babaheidarian2018feature}. Full-spectral and one-step methods model each channel as polychromatic, which requires the source spectrum and detector response but yields automatic beam-hardening correction and direct material decomposition~\cite{https://doi.org/10.1118/1.4820478, Weidinger2016PISMIR, Long2014, Kazantsev_2018}, often with a discretized energy integral for efficiency~\cite{FoygelBarber_2016, Tilley_2019, ehn2017basis}. These methods are accurate but computationally expensive and sensitive to the choice of regularizer~\cite{Bousse_2024}.
\noindent\textbf{Learning-based approaches.}
Learning-based methods address these limitations by replacing hand-crafted priors with learned ones, and differ mainly in how much physics they retain. Physics-based variants regress basis material images with a U-Net~\cite{https://doi.org/10.1002/mp.13489} or learn corrections to the polychromatic forward model~\cite{cong2018monochromaticctimagereconstruction}. Unrolled networks map each iteration of an MBIR solver to a network block and learn thresholds, step sizes or denoisers~\cite{chen2022soulnetsparselowrankunrolling, 10906250, 10475530, ge2023mbdectnetmodelbasedunrollednetwork, wang2024endtoendmodelbaseddeeplearning}, while plug-and-play methods keep the iterative loop but replace the prior with a trained denoiser or diffusion model~\cite{guo2023spectral2spectralimagespectralsimilarityassisted, jiang2024ctmaterialdecompositionusing, Jiang_2025, volumetric_dps, vazia2024spectralcttwosteponestep}. Purely data-driven methods drop the physical model entirely and rely on U-Net~\cite{doi:10.3233/XST-190500, https://doi.org/10.1002/mp.14523, 9343830, WU2021342dlbased_spectralCT, mustafa2021sparseviewspectralctreconstruction, lee2021_ultraspectralCT} or GAN architectures~\cite{https://doi.org/10.1002/mp.14828, WANG2022105952, Guo2023, QIMS76581}. All are fast at inference, but need large training sets, usually operate in the image domain, typically train one model per energy channel, and can hallucinate structures on out-of-distribution data~\cite{bhadra2021hallucinationstomographicimagereconstruction}.
Recently, multi-view scene representations have achieved state-of-the-art results by optimizing a single scene from its own projections, without external training data, using either an implicit (e.g. NeRF~\cite{mildenhall2020nerfrepresentingscenesneural}) or an explicit (e.g. 3DGS~\cite{kerbl20233dgaussiansplattingrealtime}) representation. NeRF-based CT reconstruction methods have emerged that optimize the field directly on the measured projections and recover accurate volumes even from sparse views~\cite{Zha_2022, rückert2022neatneuraladaptivetomography, 9710599, cai2024structureawaresparseviewxray3d, coronafigueroa2022mednerfmedicalneuralradiance}. Spectral extensions add basis material physics to the field~\cite{shi2024raydrivenspectralctreconstruction, xie2024acindsparsectreconstruction, Hotta2025SparseNeRF}, decompose in the projection domain~\cite{10.1117/12.3047354}, or query attenuation as a function of position and energy~\cite{10794200}. As the implicit field must be sampled densely along every ray the training and rendering remain slow. 3DGS has also been extended to CT reconstruction, and these methods overcome the sampling bottleneck by rendering an explicit set of Gaussians through differentiable rasterization instead of querying a network along every ray, which makes training and inference considerably faster. X-Gaussian~\cite{cai2024radiativegaussiansplattingefficient}, GaSpCT~\cite{nikolakakis2024gaspctgaussiansplattingnovel} and DIF-Gaussian~\cite{lin2024learning3dgaussiansextremely} adapt the representation by replacing color with attenuation, while DDGS-CT~\cite{gao2024ddgsctdirectiondisentangledgaussiansplatting} and R2-Gaussian~\cite{r2_gaussian} model X-ray physics explicitly and add a voxelizer that recovers a consistent 3D volume; further work improves efficiency and sparse-view quality~\cite{wu2024discretizedgaussianrepresentationtomographic, liu2025xgrmlargegaussianreconstruction, LI2025103585}. Spectral 3DGS variants exist in optical rendering~\cite{sinha2024spectralgaussianssemanticspectral3d, thirgood2024hypergshyperspectral3dgaussian}, but model view-dependent appearance rather than energy-dependent attenuation. Since no Gaussian representation is spectral or supports direct material decomposition, we introduce a spectral Gaussian representation that jointly enables novel view synthesis, volume reconstruction and basis material decomposition in one projection-domain optimization.
\section{Methodology}
\label{sec:methodololgy}
In this section we present our spectral CT reconstruction method (Fig.~\ref{fig:pipeline_spectral_ct_gaussians}), which represents the scanned object as 3D Gaussians carrying geometry (position, orientation, scale), an attenuation density, and basis material coefficients encoding photoelectric, Compton, and K-edge contributions, coupled to a physics-based spectral forward model. We explain each step in the following subsections and for detailed derivations refer to the supplementary material.
\subsection{Spectral CT Gaussian Representation}
We represent the scanned object as a set of $N$ explicit, differentiable 3D Gaussians~\cite{kerbl20233dgaussiansplattingrealtime}, each centered at $\mathbf{p}_i$ with covariance $\mathbf{\Sigma}_i$ parameterized by a scale vector $\mathbf{s}_i$ and a rotation quaternion $\mathbf{q}_i$. Unlike 3DGS, where spherical harmonics encode a view-dependent appearance, CT reconstruction recovers a physical attenuation field. Each Gaussian therefore carries a scalar density magnitude $\rho_i$, and its attenuation contribution at a world-space point $\mathbf{x} \in \mathbb{R}^3$ is
\begin{equation}
    G_i^3(\mathbf{x}\,|\,\rho_i, \mathbf{p}_i, \mathbf{\Sigma}_i)
    = \rho_i \exp \Bigl(-\tfrac{1}{2}(\mathbf{x} - \mathbf{p}_i)^\top \mathbf{\Sigma}_i^{-1} (\mathbf{x} - \mathbf{p}_i) \Bigr),
    \label{eq:gaussian}
\end{equation}
and the overall attenuation is accumulated additively as $\mu(\mathbf{x}) = \sum_{i=1}^{N} G_i^3(\mathbf{x})$~\cite{r2_gaussian, nikolakakis2024gaspctgaussiansplattingnovel, cai2024radiativegaussiansplattingefficient}. In spectral CT, attenuation additionally depends on photon energy $E$. Following the basis material decomposition of Alvarez and Macovski~\cite{alvarez1976_energySelectiveCT}, we write $\mu(\mathbf{x},E)=\sum_{m=1}^{M}\widetilde{\mu}_m(E)\,\alpha_m(\mathbf{x})$, where the basis functions $\widetilde{\mu}_m(E)$ encode fixed energy-dependent attenuation profiles (e.g., photoelectric, Compton, or K-edge behavior) and are shared across all Gaussians, while the spatial coefficients $\alpha_m(\mathbf{x})$ are accumulated from per-Gaussian basis coefficients $b_{m,i}$, analogously to Eq.~\ref{eq:gaussian}. To decouple geometry from material composition and to retain standard density-based pruning, we factorize $b_{m,i}=\rho_i f_{m,i}$ with a shared density $\rho_i \geq 0$ and material functions $f_{m,i} \geq 0$ satisfying $\sum_{m=1}^{M} f_{m,i} = 1$~\cite{So_spectralCTPrinciples,Long2014,Ding_2018,liu2022dual}, resulting in
\begin{equation}
    \displaystyle
        \mu(\mathbf{x}, E)
        = \sum_{m=1}^{M} \widetilde{\mu}_m(E)
        \Biggl( \sum_{i=1}^{N} \rho_i f_{m,i} \exp \Bigl(-\tfrac{1}{2}
        (\mathbf{x} - \mathbf{p}_i)^\top \mathbf{\Sigma}_i^{-1} 
        (\mathbf{x} - \mathbf{p}_i) \Bigr) \Biggr).
    \label{eq:constrained_basis_material_model}
\end{equation}
\subsection{Spectral CT Rendering}
Spectral CT requires a rendering model that accounts for the energy-dependent attenuation of the object and the spectral response of each detector channel. Let $\mathbf{r}(s)$ denote a ray cast from the source through the object, parameterized by the arc length $s$ and bounded by the near and far intersections $s_n$ and $s_f$ with the reconstruction volume. The predicted intensity of ray $\mathbf{r}$ in channel $c$ is then modeled with the polychromatic Beer--Lambert law
\begin{equation}
    \displaystyle
    \widehat{I}_c(\mathbf{r})
    = \int_{E_{\min}}^{E_{\max}}
    I_0(E)\,S_c(E)
    \exp \Bigl(- \int_{s_n}^{s_f} \mu(\mathbf{r}(s),E)\,ds \Bigr)\,dE,
    \label{eq:beer_lambert_continuous}
\end{equation}
where $I_0(E)$ is the source spectrum and $S_c(E)$ is the spectral sensitivity of channel $c$.. We substitute the basis material model (Eq.~\ref{eq:constrained_basis_material_model}) for $\mu(\mathbf{r}(s),E)$ and exploit the linearity of the line integral over the spatial dimension, which separates the energy dependence from the spatial one. Since the continuous energy integral is computationally infeasible, we discretize it into $K$ energy samples~\cite{cong2018monochromaticctimagereconstruction, zhao2022statisticaliterativespectralct, shi2024raydrivenspectralctreconstruction, volumetric_dps}, i.e.
\begin{equation}
    \displaystyle
    \widehat{I}_c(\mathbf{r})
    = \sum_{k=1}^{K}
    I_0(E_k)\,S_c(E_k)\,w_{k}
    \exp \Bigl(-
    \sum_{m=1}^{M} \widetilde{\mu}_m(E_k)
    \int_{s_n}^{s_f} \alpha_{m}(\mathbf{r}(s))\,ds
    \Bigr),
    \label{eq:beer_lambert_spectral_discrete}
\end{equation}
where $(E_k, w_k)$ is the energy grid obtained by discretizing the interval $[20,140]~\mathrm{keV}$ into $K=64$ uniform midpoint samples. To compare against the observed log-transformed projections, we apply the log-transform $\widehat{y}_c(\mathbf{r}) = -\log(\widehat{I}_c(\mathbf{r})/I_{0,c})$. The channel incident intensities $I_{0,c}$ are estimated from flat-field scans or air patches in the raw projections. During rendering, the Gaussians are first rasterized into projected basis-coefficient images, an energy-agnostic step computed once per iteration. The spectral model then combines these projections across the energy grid into the final measurements. The two components, the rasterizer and the basis functions $\widetilde{\mu}_m(E)$, are described next.
\noindent\textbf{Rasterization}
The rasterizer computes the projected basis-coefficient line integrals $L_m(\mathbf{r})$ required by Eq.~\eqref{eq:beer_lambert_spectral_discrete}. We extend the R2-Gaussian rasterizer~\cite{r2_gaussian} to project $M$ basis channels instead of a single density. Each Gaussian is mapped onto the detector plane, yielding a projected mean $\widehat{\mathbf{p}}_i$, covariance $\widehat{\mathbf{\Sigma}}_i$, and rescaled density $\widehat{\rho}_i$~\cite{r2_gaussian}. Since the resulting 2D kernel is shared across all $M$ bases and only the scalar material function $f_{m,i}$ varies, the projection at detector pixel $\widehat{\mathbf{x}}_{u,v}$ is
\begin{equation}
    \displaystyle
    L_{m}(\widehat{\mathbf{x}}_{u,v}) \approx \sum_{i=1}^{N} 
    \widehat{\rho}_{i}\, f_{m,i}
    \exp\Bigl( -\tfrac{1}{2} (\widehat{\mathbf{x}}_{u,v}-\widehat{\mathbf{p}}_i)^\top 
    \widehat{\mathbf{\Sigma}}_i^{-1}
    (\widehat{\mathbf{x}}_{u,v}-\widehat{\mathbf{p}}_i) \Bigr),
    \label{eq:rasterized_basis}
\end{equation}
giving a multi-channel projection image of shape $(M, H, W)$ for detector dimensions $(H, W)$.
\noindent\textbf{Basis functions}
The basis functions $\widetilde{\mu}_m(E)$ are collected into an energy-basis matrix $\mathbf{B}_E \in \mathbb{R}^{K \times M}$ with $\mathbf{B}_E[k,m] = \widetilde{\mu}_m(E_k)$. We use the physics-derived photoelectric and Compton bases of Alvarez and Macovski~\cite{alvarez1976_energySelectiveCT}, where $\widetilde{\mu}_{\mathrm{PE}}(E) = E^{-3}$ dominates at low energies and is sensitive to high-$Z$ materials such as bone, while the Klein--Nishina Compton basis $\widetilde{\mu}_{\mathrm{C}}(E)$ captures the energy dependence at higher energies and is more sensitive to soft tissue. Since this two-basis model cannot represent the sharp attenuation jump of K-edge materials, we optionally add a learnable third basis modeled as a sigmoid centered at a learned edge energy $E_{\mathrm{K}}$
\begin{equation}
    \widetilde{\mu}_{\mathrm{K}}(E)
    = \alpha_{\mathrm{K}}\,\sigma\!\left(\gamma_{\mathrm{K}}
    (E - E_{\mathrm{K}})\right),
\end{equation}
where $\alpha_{\mathrm{K}}$ controls the amplitude and $\gamma_{\mathrm{K}}$ the sharpness. These parameters are optimized jointly with the Gaussian model and initialized for iodine, a commonly used clinical contrast agent.

\noindent\textbf{Basis conditioning}
The raw K-edge curve partly overlaps with the photoelectric and Compton curves over the sampled energy range, so the model can represent the same measurement by shifting weight between basis coefficients, which leads to ambiguous solutions and unstable gradients~\cite{Clark2017HybridSpectralCT, Zeegers_2022, Wu2020DLIMD, rahman2023directiterativereconstructionmultiple}. We therefore orthogonalize the K-edge column against the photoelectric and Compton columns under a diagonal weighting that emphasizes energy samples contributing more to the measured signal, which lets the learned K-edge coefficient focus on the residual edge-like response. All columns are normalized at a reference energy $E_{\mathrm{ref}} = 70\,\mathrm{keV}$, which removes their arbitrary scale without changing the span of the spectral model.
\definecolor{turboA}{rgb}{0.190,0.072,0.232}
\definecolor{turboB}{rgb}{0.244,0.288,0.538}
\definecolor{turboC}{rgb}{0.255,0.496,0.803}
\definecolor{turboD}{rgb}{0.212,0.665,0.925}
\definecolor{turboE}{rgb}{0.164,0.817,0.725}
\definecolor{turboF}{rgb}{0.311,0.924,0.506}
\definecolor{turboG}{rgb}{0.591,0.981,0.314}
\definecolor{turboH}{rgb}{0.826,0.896,0.185}
\definecolor{turboI}{rgb}{0.968,0.706,0.123}
\definecolor{turboJ}{rgb}{0.944,0.447,0.115}
\definecolor{turboK}{rgb}{0.740,0.214,0.251}
\pgfdeclareverticalshading{errorcolorbarshading}{100bp}{
    color(0bp)=(turboA);
    color(10bp)=(turboB);
    color(20bp)=(turboC);
    color(30bp)=(turboD);
    color(40bp)=(turboE);
    color(50bp)=(turboF);
    color(60bp)=(turboG);
    color(70bp)=(turboH);
    color(80bp)=(turboI);
    color(90bp)=(turboJ);
    color(100bp)=(turboK)
}
\pgfdeclarehorizontalshading{errorcolorbarshadinghoriz}{100bp}{
    color(0bp)=(turboA);
    color(10bp)=(turboB);
    color(20bp)=(turboC);
    color(30bp)=(turboD);
    color(40bp)=(turboE);
    color(50bp)=(turboF);
    color(60bp)=(turboG);
    color(70bp)=(turboH);
    color(80bp)=(turboI);
    color(90bp)=(turboJ);
    color(100bp)=(turboK)
}
\newcommand{\errorcolorbar}[1][6.5cm]{%
    \begin{tikzpicture}[x=1cm,y=#1]
        \shade[shading=errorcolorbarshading] (0,0) rectangle (0.22,1);
        \draw[black, line width=0.2pt] (0,0) rectangle (0.22,1);
        \draw[black, line width=0.2pt] (0.22,0.0) -- (0.32,0.0);
        \draw[black, line width=0.2pt] (0.22,0.5) -- (0.32,0.5);
        \draw[black, line width=0.2pt] (0.22,1.0) -- (0.32,1.0);
        \node[anchor=west, font=\scriptsize] at (0.36,0.0) {0};
        \node[anchor=west, font=\scriptsize] at (0.36,0.5) {0.5};
        \node[anchor=west, font=\scriptsize] at (0.36,1.0) {1};
    \end{tikzpicture}%
}
\newcommand{\errorcolorbarhoriz}[1][4cm]{%
    \begin{tikzpicture}[x=#1,y=1cm]
        \shade[shading=errorcolorbarshadinghoriz] (0,0) rectangle (1,0.22);
        \draw[black, line width=0.2pt] (0,0) rectangle (1,0.22);
        \draw[black, line width=0.2pt] (0.0,0.22) -- (0.0,0.32);
        \draw[black, line width=0.2pt] (0.5,0.22) -- (0.5,0.32);
        \draw[black, line width=0.2pt] (1.0,0.22) -- (1.0,0.32);
        \node[anchor=south, font=\scriptsize] at (0.0,0.36) {0};
        \node[anchor=south, font=\scriptsize] at (0.5,0.36) {0.5};
        \node[anchor=south, font=\scriptsize] at (1.0,0.36) {1};
    \end{tikzpicture}%
}
\subsection{Training}
\noindent\textbf{Initialization}
We initialize the Gaussian positions from a pseudo-ground-truth FDK volume, as in R2-Gaussian~\cite{r2_gaussian}. Since FDK is monochromatic, we reconstruct from the lowest energy channel only, where the enhanced photoelectric contrast provides clearer boundaries for the initial geometry~\cite{So_spectralCTPrinciples, doi:10.1148/radiol.2015142631, GREFFIER2023167}. Beyond geometry, we also initialize the basis material functions from the same volume, which R2-Gaussian~\cite{r2_gaussian} does not address. A physics-motivated heuristic derives the functions from the log-density, biasing denser Gaussians toward the photoelectric basis and less dense ones toward Compton, $f_{\mathrm{PE},i} = \sigma(a(\log\rho_i - \log\rho_0))$ and $f_{\mathrm{C},i} = 1 - f_{\mathrm{PE},i}$. For the three-basis model a density-dependent gate $g_i = \sigma(a_{\mathrm{K}}(\rho_i/\rho_{\mathrm{K}}-1))$ reassigns part of this photoelectric budget to the K-edge basis, $f_{\mathrm{K},i} = f_{\mathrm{K},\max}\,g_i\,f_{\mathrm{PE},i}$, where $\rho_0$, $\rho_{\mathrm{K}}$ and the slopes are set from quantiles of the initial density distribution. This only avoids a symmetric initialization of the functions.
\noindent\textbf{Optimization}
We optimize the Gaussian parameters with Adam~\cite{kingma2017adammethodstochasticoptimization} using separate learning-rate schedules per parameter group. The material functions receive a higher learning rate than the density, which encourages the model to explain attenuation variation through material composition rather than density alone. In the three-basis model the K-edge parameters are frozen during a warm-up period, so that geometry and functions can stabilize before the K-edge component is introduced. For each energy channel $c$ we compare the rendered log-projection $\widehat{\mathbf{y}}_c$ against the measurement $\mathbf{y}_c$ with a combination of L1 and D-SSIM losses~\cite{kerbl20233dgaussiansplattingrealtime} and a 3D total-variation term following R2-Gaussian~\cite{r2_gaussian}, and add a 2D total-variation term on each rasterized basis-coefficient image $\mathbf{L}_m$ to promote spatially smooth decompositions as represented below:
\begin{align}
    \mathcal{L}
    = \sum_{c=1}^{C} \Bigl[
    &(1-\lambda_{\mathrm{S}})\,\mathcal{L}_{1}(\widehat{\mathbf{y}}_c, \mathbf{y}_c)
    + \lambda_{\mathrm{S}}\,\mathcal{L}_{\mathrm{D\text{-}SSIM}}(\widehat{\mathbf{y}}_c, \mathbf{y}_c) \nonumber \\
    &+ \tfrac{\lambda_{\mathrm{3D}}}{C}\,\mathcal{L}_{\mathrm{3DTV}}(\mathbf{V}_{\mathrm{TV},c}) \Bigr]
    + \sum_{m=1}^{M} \lambda_{m}\,\mathcal{L}_{\mathrm{TV}}(\mathbf{L}_{m}).
    \label{eq:total_loss}
\end{align}
We further extend the R2-Gaussian densification and pruning strategy~\cite{r2_gaussian} with a separate split-gradient threshold and reset the basis material functions at each opacity-reset interval to prevent numerical inactivity.
\begin{table*}[htbp!]
\centering
\begin{minipage}[t]{0.62\textwidth}
\centering
\caption{Quantitative results on the multi-energy bird chest dataset~\cite{BIRDCHEST_meaney_2024_11388131}.}
\label{tab:quail-metrics}
\resizebox{\linewidth}{!}{%
    \setlength{\tabcolsep}{4pt}
    \begin{tabular}{lccc ccc ccc ccc}
    \toprule
    \multirow{2}{*}{Method}
    & \multicolumn{3}{c}{50 keV}
    & \multicolumn{3}{c}{80 keV}
    & \multicolumn{3}{c}{120 keV}
    & \multicolumn{3}{c}{Average} \\
    \cmidrule(lr){2-4}\cmidrule(lr){5-7}\cmidrule(lr){8-10}\cmidrule(lr){11-13}
    & PSNR$\uparrow$ & SSIM$\uparrow$ & LPIPS$\downarrow$
    & PSNR$\uparrow$ & SSIM$\uparrow$ & LPIPS$\downarrow$
    & PSNR$\uparrow$ & SSIM$\uparrow$ & LPIPS$\downarrow$
    & PSNR$\uparrow$ & SSIM$\uparrow$ & LPIPS$\downarrow$ \\
    \midrule
    \multicolumn{13}{l}{\textit{Novel view synthesis}} \\
    \midrule
    FDK \cite{Feldkamp:84}                                        & 27.86 & 0.627 & \textbf{0.208} & 30.28 & 0.669 & \textbf{0.171} & 31.43 & 0.683 & \textbf{0.182} & 29.85 & 0.660 & \textbf{0.187} \\
    SART \cite{ANDERSEN198481}                                    & 31.45 & 0.838 & 0.262 & 34.40 & 0.869 & 0.233 & 37.24 & 0.883 & 0.255 & 34.36 & 0.863 & 0.250 \\
    CGLS \cite{Soleimani2015IterativeCT}                          & 31.19 & 0.785 & 0.244 & 34.14 & 0.833 & 0.216 & 37.13 & 0.872 & 0.242 & 34.15 & 0.830 & 0.234 \\
    PDHG-TV \cite{Chambolle2011}                                  & 31.75 & 0.824 & 0.269 & 34.34 & 0.855 & 0.237 & 36.71 & 0.876 & 0.258 & 34.27 & 0.852 & 0.255 \\
    \midrule
    IntraTomo \cite{9710599}                                      & 29.81 & 0.830 & 0.330 & 30.83 & 0.833 & 0.453 & 23.80 & 0.746 & 0.415 & 28.15 & 0.803 & 0.399 \\
    SAX-NeRF \cite{cai2024structureawaresparseviewxray3d}         & 33.58 & 0.906 & 0.271 & 35.08 & 0.887 & 0.380 & 25.63 & 0.769 & 0.343 & 31.43 & 0.854 & 0.332 \\
    X-Gaussian \cite{cai2024radiativegaussiansplattingefficient}  & \textbf{34.66} & 0.931 & 0.220 & 21.14 & 0.866 & 0.194 & 20.97 & 0.764 & 0.303 & 25.59 & 0.854 & 0.239 \\
    R2-Gaussian \cite{r2_gaussian}                                & 34.07 & \textbf{0.939} & 0.274 & \textbf{37.43} & \textbf{0.958} & 0.248 & \textbf{40.50} & \textbf{0.960} & 0.294 & \textbf{37.33} & \textbf{0.953} & 0.272 \\
    X-Field \cite{wang2025xfieldphysicallygroundedrepresentation} & 27.31 & 0.866 & 0.298 & 26.99 & 0.854 & 0.415 & 32.22 & 0.897 & 0.331 & 28.84 & 0.872 & 0.348 \\
    \midrule
    Ours (2)                                                      & 32.99 & 0.929 & 0.285 & 35.67 & 0.945 & 0.255 & 39.09 & 0.945 & 0.312 & 35.92 & 0.940 & 0.284 \\
    Ours (3)                                                      & 33.76 & 0.930 & 0.282 & 36.79 & 0.949 & 0.254 & 39.03 & 0.949 & 0.313 & 36.52 & 0.943 & 0.283 \\
    \midrule
    \multicolumn{13}{l}{\textit{Spectral CT reconstruction}} \\
    \midrule
    FDK \cite{Feldkamp:84}                                        & 19.40 & 0.383 & \textbf{0.380} & 21.94 & 0.348 & \textbf{0.366} & 18.92 & 0.227 & \textbf{0.374} & 20.09 & 0.319 & \textbf{0.373} \\
    SART \cite{ANDERSEN198481}                                    & 22.68 & 0.650 & 0.415 & 25.60 & 0.621 & 0.417 & 25.20 & 0.390 & 0.456 & 24.49 & 0.554 & 0.429 \\
    CGLS \cite{Soleimani2015IterativeCT}                          & 18.56 & 0.346 & 0.463 & 19.95 & 0.314 & 0.452 & 17.36 & 0.149 & 0.455 & 18.62 & 0.269 & 0.457 \\
    PDHG-TV \cite{Chambolle2011}                                  & 22.25 & \textbf{0.687} & 0.426 & 24.75 & 0.675 & 0.446 & 23.83 & 0.435 & 0.559 & 23.61 & 0.599 & 0.477 \\
    \midrule
    IntraTomo \cite{9710599}                                      & 20.03 & 0.537 & 0.416 & 24.24 & 0.567 & 0.413 & 27.03 & 0.591 & 0.405 & 23.77 & 0.565 & 0.411 \\
    SAX-NeRF \cite{cai2024structureawaresparseviewxray3d}         & 17.76 & 0.492 & 0.468 & 20.87 & 0.482 & 0.469 & 24.04 & 0.534 & 0.446 & 20.89 & 0.503 & 0.461 \\
    R2-Gaussian \cite{r2_gaussian}                                & 20.06 & 0.631 & 0.405 & 24.43 & \textbf{0.678} & 0.393 & \textbf{27.67} & \textbf{0.715} & 0.387 & 24.06 & \textbf{0.674} & 0.395 \\
    \midrule
    Ours (2)                                                      & \textbf{26.54} & 0.610 & 0.404 & \textbf{27.19} & 0.616 & 0.395 & 27.09 & 0.609 & 0.389 & \textbf{26.94} & 0.612 & 0.396 \\
    Ours (3)                                                      & 25.64 & 0.627 & 0.409 & 27.02 & 0.632 & 0.400 & 27.05 & 0.644 & 0.399 & 26.57 & 0.634 & 0.402 \\
    \bottomrule
    \end{tabular}}
\end{minipage}\hfill
\begin{minipage}[t]{0.37\textwidth}
\centering
\caption{Basis volume reconstruction (photoelectric, Compton, and K-edge) quantitative results on the synthetic dataset.}
\label{tab:bmd-synthetic3basis-metrics}
\resizebox{\linewidth}{!}{%
    \setlength{\tabcolsep}{1pt}
    \begin{tabular}{l cc cc cc cc}
    \toprule
    \multirow{2}{*}{Method}
    & \multicolumn{2}{c}{PE}
    & \multicolumn{2}{c}{Compton}
    & \multicolumn{2}{c}{K-edge}
    & \multicolumn{2}{c}{Average} \\
    \cmidrule(lr){2-3}\cmidrule(lr){4-5}\cmidrule(lr){6-7}\cmidrule(lr){8-9}
    & PSNR$\uparrow$ & SSIM$\uparrow$
    & PSNR$\uparrow$ & SSIM$\uparrow$
    & PSNR$\uparrow$ & SSIM$\uparrow$
    & PSNR$\uparrow$ & SSIM$\uparrow$ \\
    \midrule
    FDK \cite{Feldkamp:84}               & 20.04 & 0.548 & 21.41 & 0.694 & 25.64 & 0.573 & 22.36 & 0.605 \\
    SART \cite{ANDERSEN198481}           & 20.81 & 0.666 & 21.44 & 0.779 & 25.90 & 0.596 & \textbf{22.72} & 0.681 \\
    CGLS \cite{Soleimani2015IterativeCT} & 20.09 & 0.684 & \textbf{21.63} & 0.806 & 25.61 & 0.606 & 22.44 & 0.699 \\
    PDHG-TV \cite{Chambolle2011}         & 20.67 & 0.720 & 21.38 & 0.808 & \textbf{26.03} & 0.597 & 22.70 & 0.708 \\
    Ours (2)                             & 23.64 & 0.853 & 15.56 & \textbf{0.814} & -- & -- & 19.60 & 0.833 \\
    Ours (3)                             & \textbf{23.82} & \textbf{0.863} & 15.28 & 0.808 & 20.59 & \textbf{0.898} & 19.89 & \textbf{0.856} \\
    \bottomrule
    \end{tabular}}

\vspace{0.6em}
\caption{Basis material decomposition quantitative results on the synthetic dataset using the segmented GT maps.}
\label{tab:bmd-synthetic3basis_rgb-metrics}
\resizebox{0.8\linewidth}{!}{%
    \setlength{\tabcolsep}{4pt}
    \begin{tabular}{l ccc}
    \toprule
    Method & PSNR$\uparrow$ & SSIM$\uparrow$ & LPIPS$\downarrow$ \\
    \midrule
    Ours (2) & 30.11 & 0.985 & 0.062 \\
    Ours (3) & \textbf{30.97} & \textbf{0.989} & \textbf{0.047} \\
    \bottomrule
    \end{tabular}}
\end{minipage}
\end{table*}
\subsection{Basis material decomposition}
\label{sec:method-bmd}
Once the model is optimized, the basis coefficients encoded in each Gaussian can be turned into
material maps. This step operates directly on the optimized Gaussians and is separate from the
reconstruction objective. We apply mean-shift clustering~\cite{comaniciu2002mean}, which identifies
modes in feature space without requiring a predefined number of clusters. For each Gaussian we form
the basis coefficients $b_m = \rho f_m$ and cluster them in the feature space
\begin{equation}
    \mathbf{z} =
    \begin{cases}
        \Bigl[\, \rho,\; \kappa \,\Bigr], & \text{two-basis},\\[4pt]
        \Bigl[\, \rho,\; \kappa,\; f_{\mathrm{K}} \,\Bigr], & \text{three-basis},
    \end{cases}
    \qquad
    \kappa = \log\!\Bigl(1 + \frac{b_{\mathrm{PE}}}{b_{\mathrm{C}}+\epsilon}\Bigr),
    \label{eq:bmd_features}
\end{equation}
where $b_{\mathrm{PE}}$ and $b_{\mathrm{C}}$ are the photoelectric and Compton coefficients, the small
constant $\epsilon$ keeps the ratio $\kappa$ finite for a vanishing Compton coefficient, and
the K-edge entry is only present in the three-basis model. The logarithm compresses the dynamic range
of $\kappa$ and all features are normalized before clustering. Clusters are ordered by $\kappa$ and
density, yielding soft tissue and bone-like groups, and in the three-basis model the cluster with the
largest K-edge coefficient is identified as the K-edge material. The clustered Gaussians are finally
assigned semantic colors and rendered into segmentation images.

\section{Datasets}
\label{sec:datasets}
We evaluate our method on one real-world multi-energy cone-beam dataset~\cite{BIRDCHEST_meaney_2024_11388131} and on
our synthetic spectral phantom, which provides known ground-truth basis material decomposition, as none of the
publicly available datasets offers this information. Our synthetic phantom is defined on a $128^3$ grid and consists of an outer soft-tissue cylinder containing a
dense bone cylinder and a K-edge cylinder with a sharp transition at $63$~keV, where every voxel stores the
ground-truth photoelectric, Compton and K-edge coefficients. Additional details on other real-world datasets are presented in the supplementary material.
\noindent\textbf{Dataset preprocessing}
The real-world datasets are converted to a common format with a folder per spectral channel and a shared metadata file holding the scanner geometry, spectral information and view splits. The raw count data are normalized by the approximate unattenuated incident intensity and then log-transformed (if not already log transformed). From each dataset we select $75$ evenly spaced training views and $100$ held-out test views, and all compared methods are trained and evaluated on the same splits.
\begin{figure*}[t]
\centering
\caption{Novel view synthesis qualitative results on the multi-energy bird chest dataset~\cite{BIRDCHEST_meaney_2024_11388131}. For each method the left column shows the rendered projection, the right column the absolute error w.r.t.\ the ground truth.}
\label{fig:nvs-quail-images}
\setlength{\tabcolsep}{0pt}
\renewcommand{\arraystretch}{0}
\renewcommand{\imgcell}[1]{\includegraphics[width=0.062\textwidth]{#1}}
\renewcommand{\methodcell}[1]{\scriptsize\bfseries #1}
\newcommand{\enercell}[1]{\raisebox{0.55cm}[0pt][0pt]{\scriptsize\bfseries #1}}
\scalebox{1}{
\begin{tabular}{@{}l@{\hspace{3pt}}c@{\hspace{0.15cm}}cc@{\hspace{0.15cm}}cc@{\hspace{0.15cm}}cc@{\hspace{0.15cm}}cc@{\hspace{0.15cm}}cc@{\hspace{0.15cm}}cc@{\hspace{0.15cm}}cc@{}}
    & \methodcell{GT}
    & \multicolumn{2}{c}{\methodcell{FDK}~{\scriptsize\cite{Feldkamp:84}}}
    & \multicolumn{2}{c}{\methodcell{CGLS}~{\scriptsize\cite{Soleimani2015IterativeCT}}}
    & \multicolumn{2}{c}{\methodcell{SAX-NeRF}~{\scriptsize\cite{cai2024structureawaresparseviewxray3d}}}
    & \multicolumn{2}{c}{\methodcell{R2-Gaussian}~{\scriptsize\cite{r2_gaussian}}}
    & \multicolumn{2}{c}{\methodcell{X-Gaussian}~{\scriptsize\cite{cai2024radiativegaussiansplattingefficient}}}
    & \multicolumn{2}{c}{\methodcell{Ours (2)}}
    & \multicolumn{2}{c}{\methodcell{Ours (3)}} \\
    \addlinespace[2pt]

    \enercell{50\,keV}
    & \imgcell{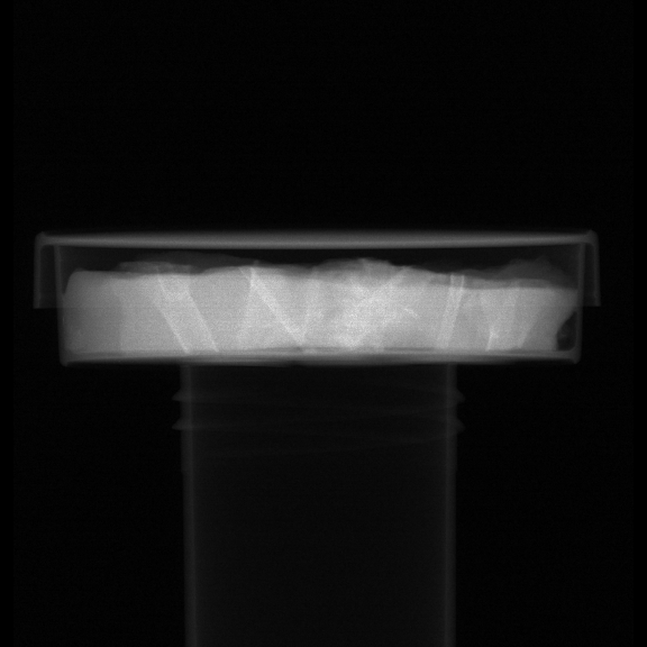}
    & \imgcell{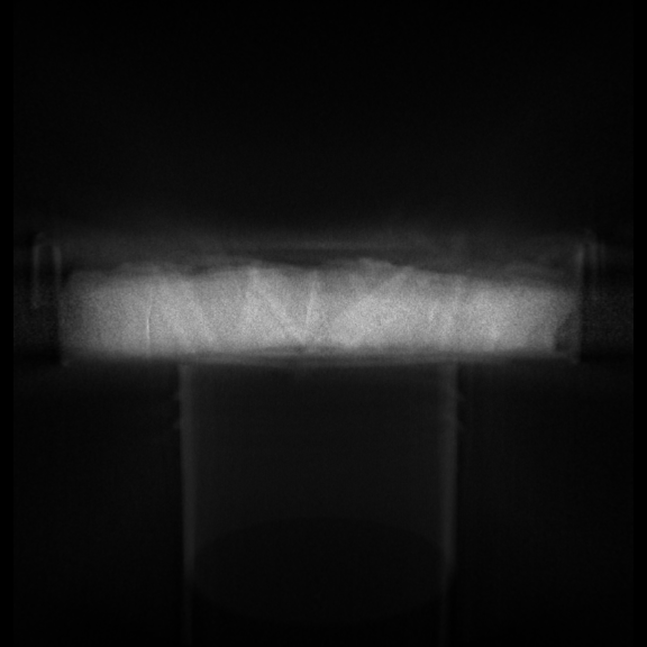}
    & \imgcell{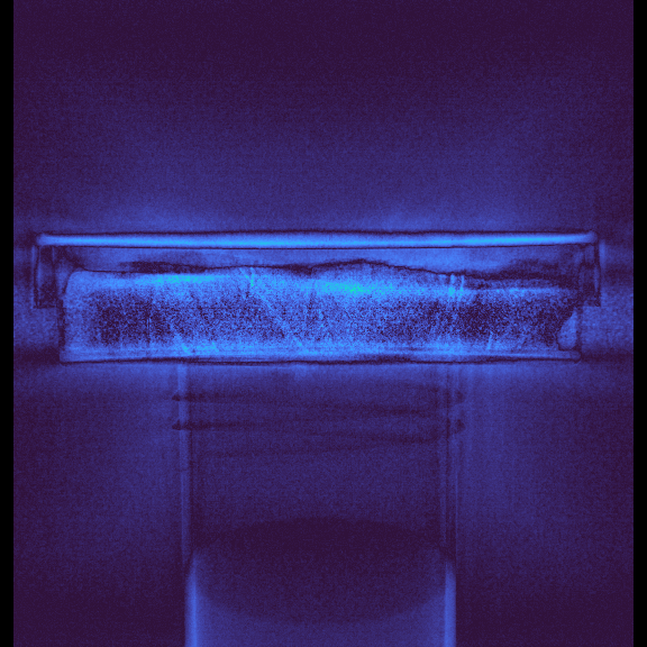}
    & \imgcell{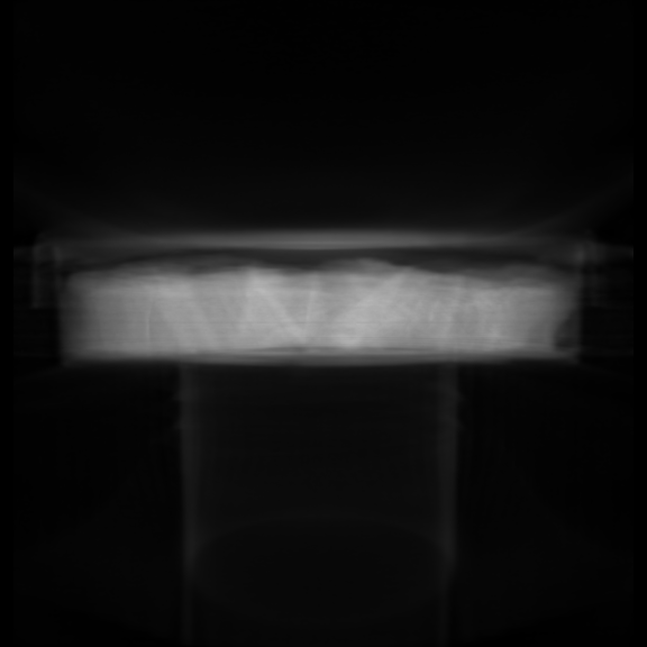}
    & \imgcell{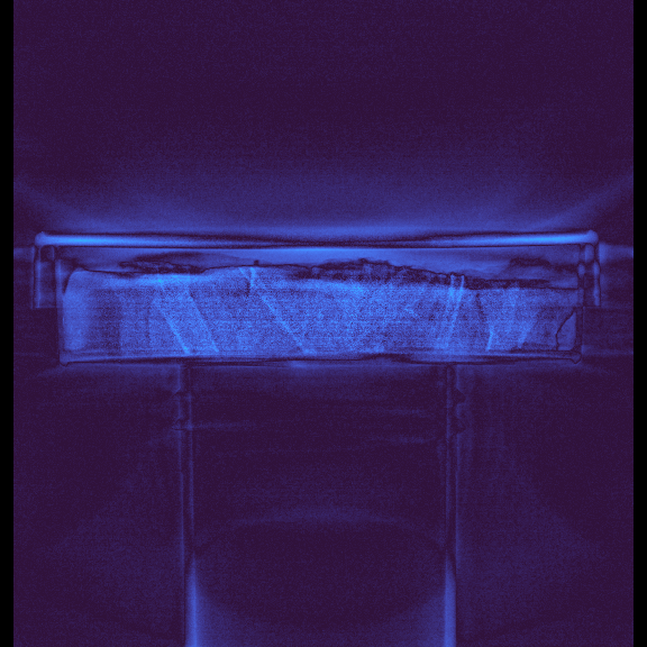}
    & \imgcell{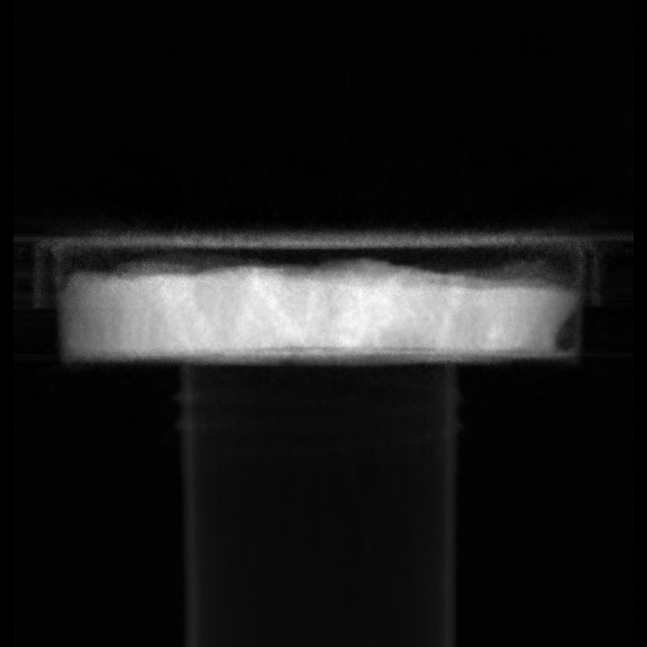}
    & \imgcell{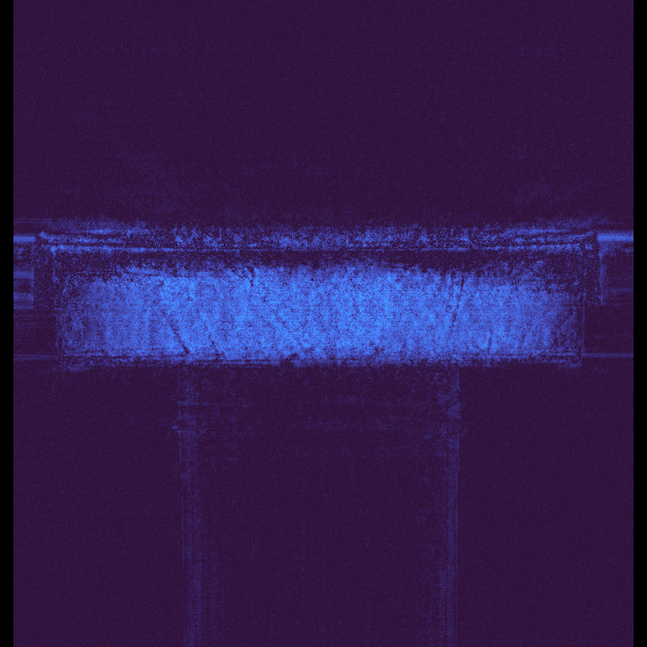}
    & \imgcell{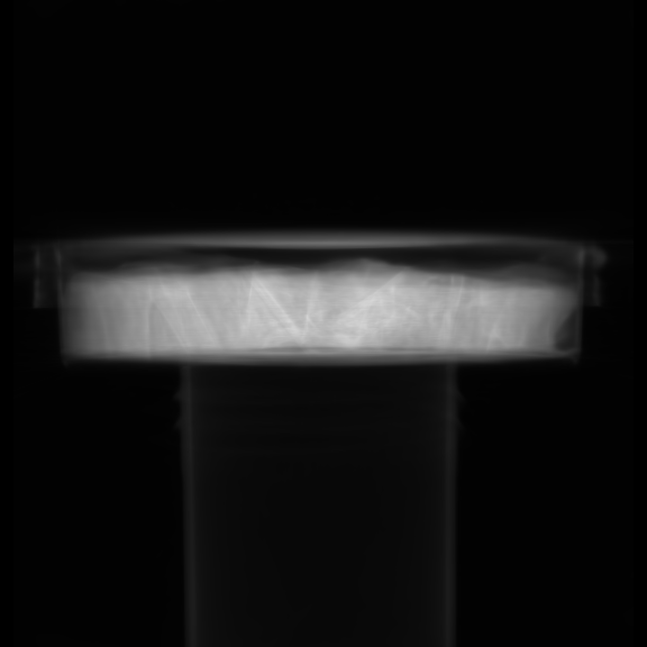}
    & \imgcell{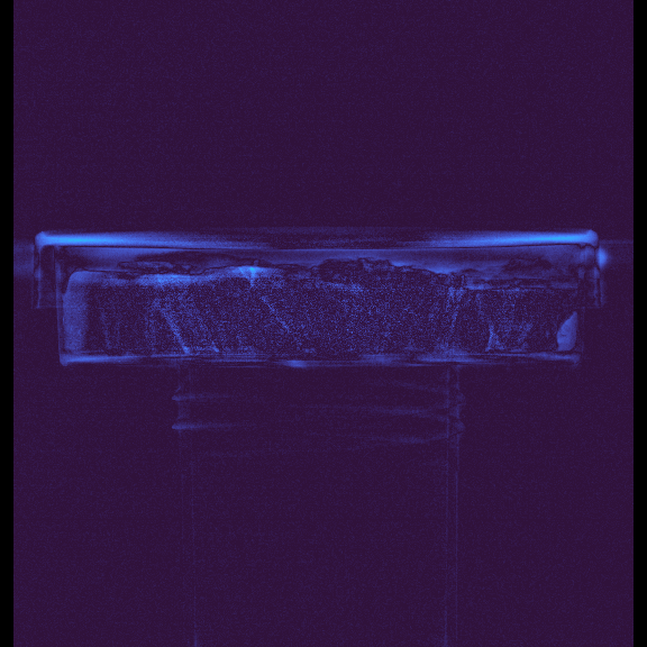}
    & \imgcell{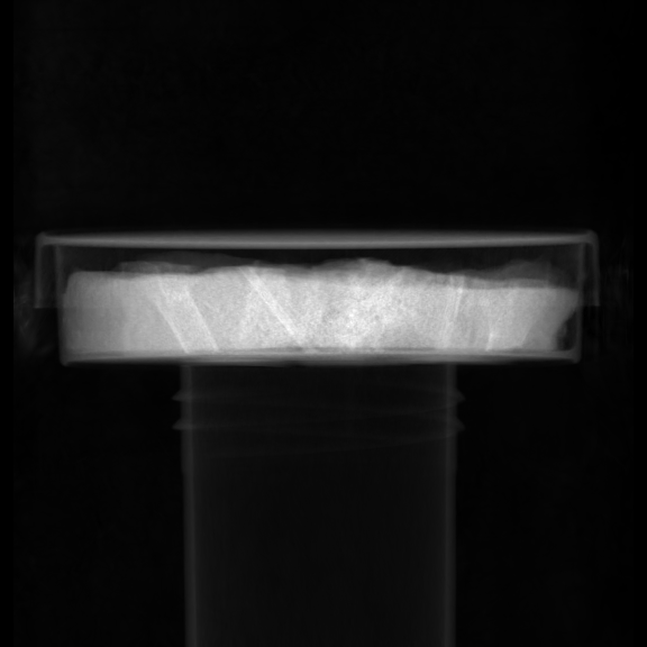}
    & \imgcell{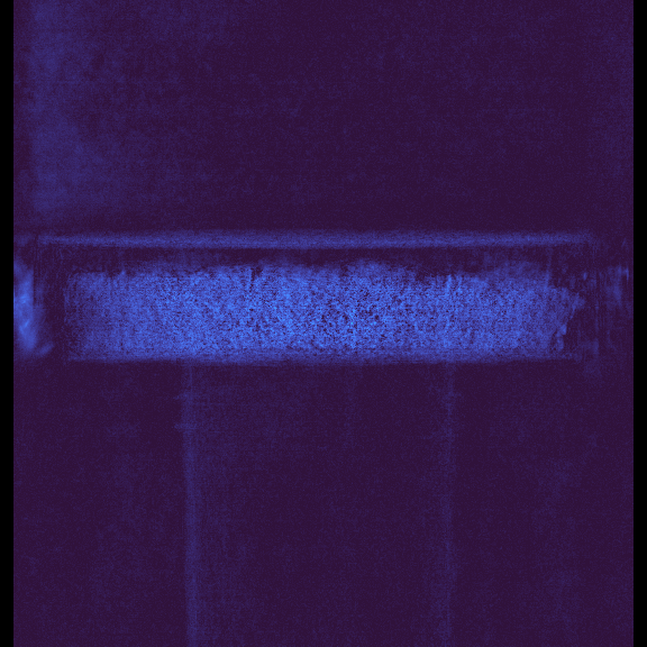}
    & \imgcell{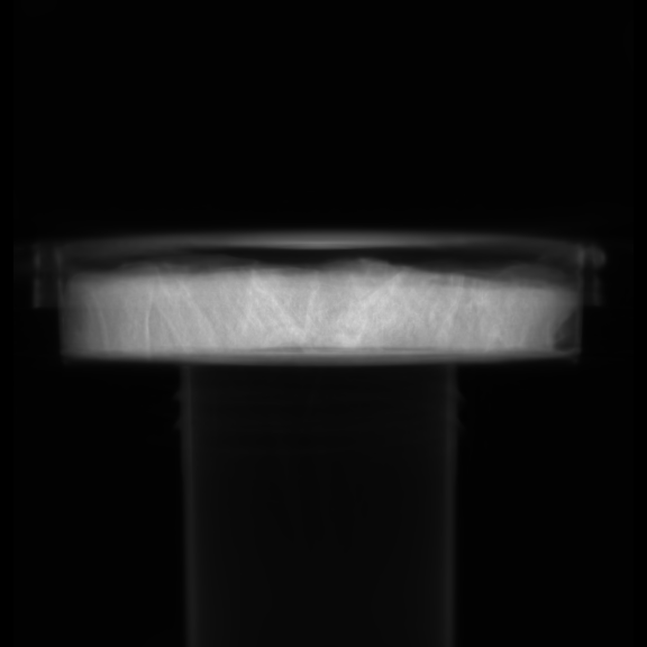}
    & \imgcell{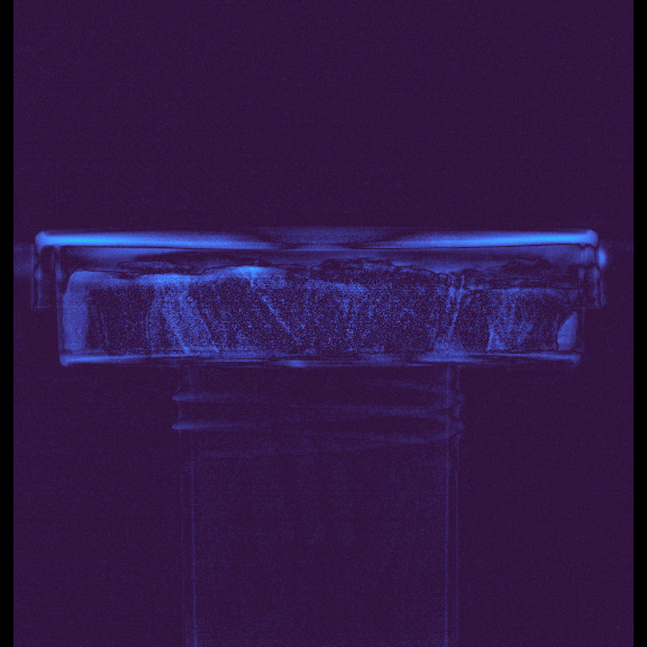}
    & \imgcell{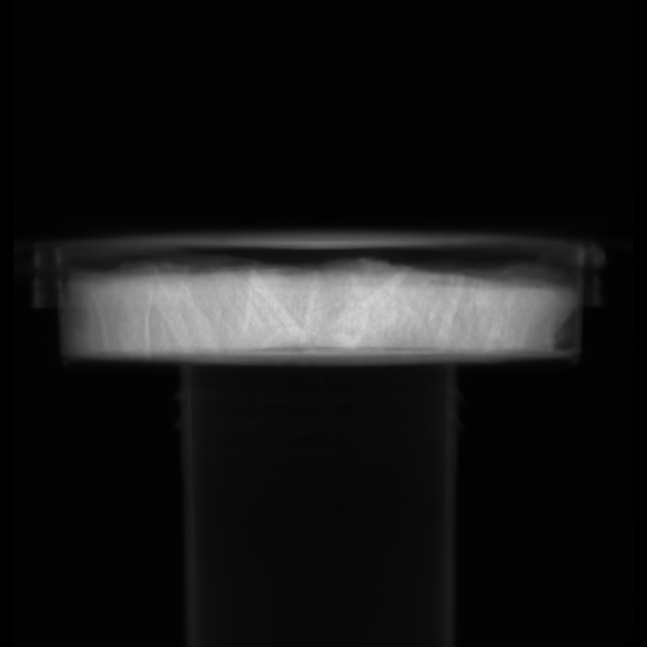}
    & \imgcell{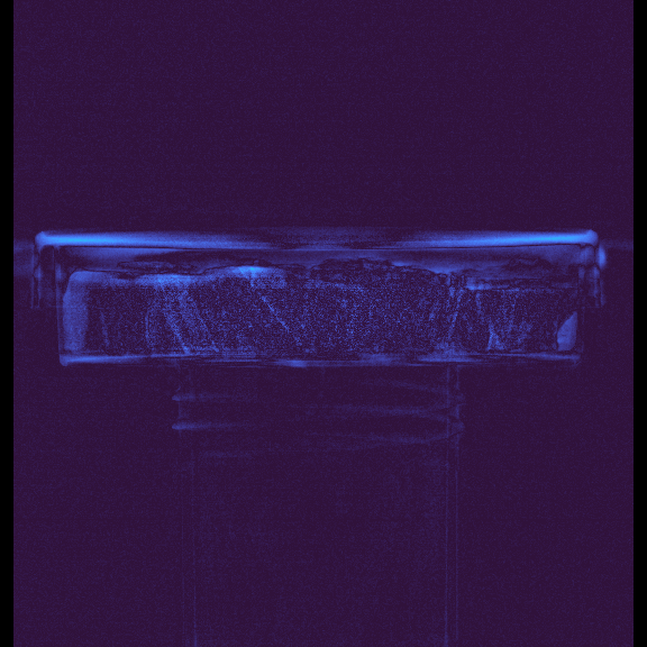}
    \\

    \enercell{80\,keV}
    & \imgcell{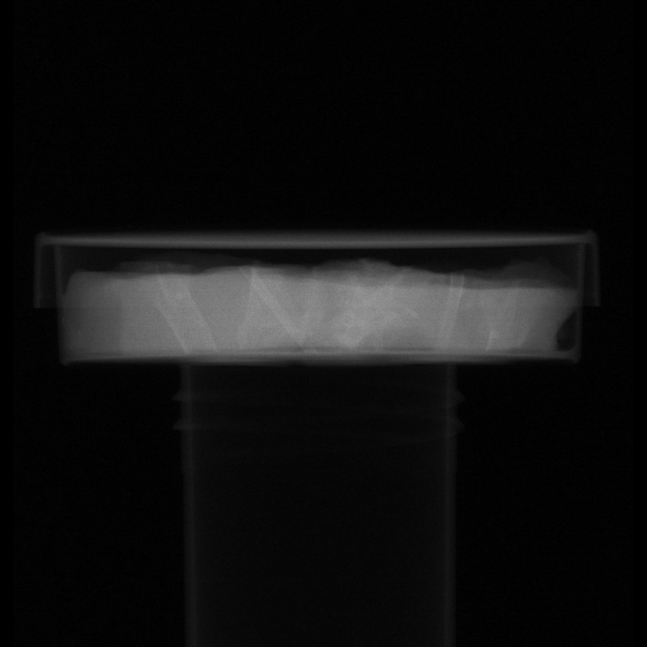}
    & \imgcell{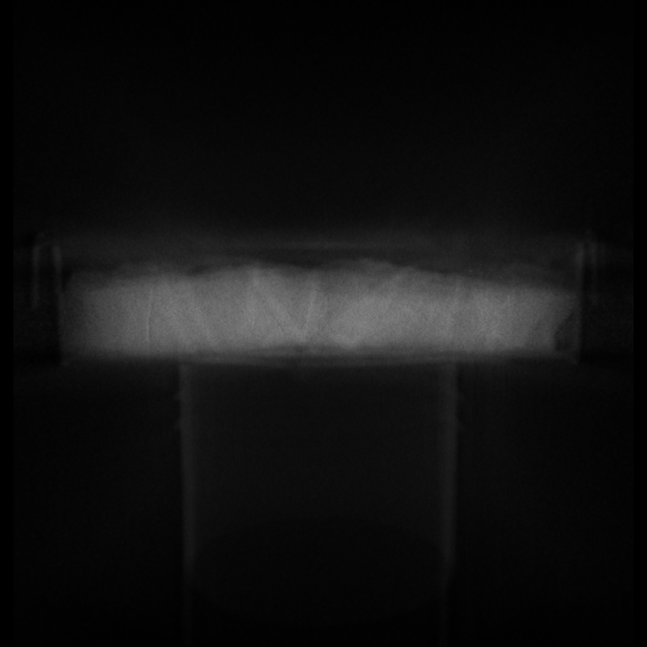}
    & \imgcell{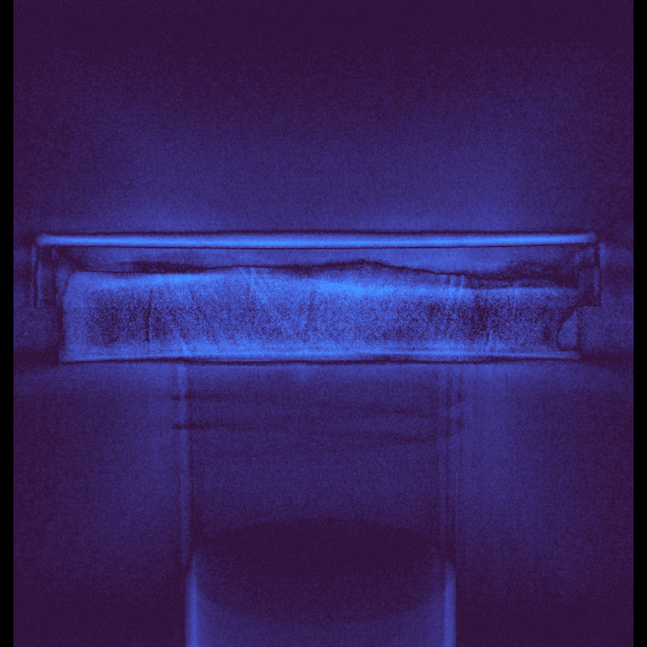}
    & \imgcell{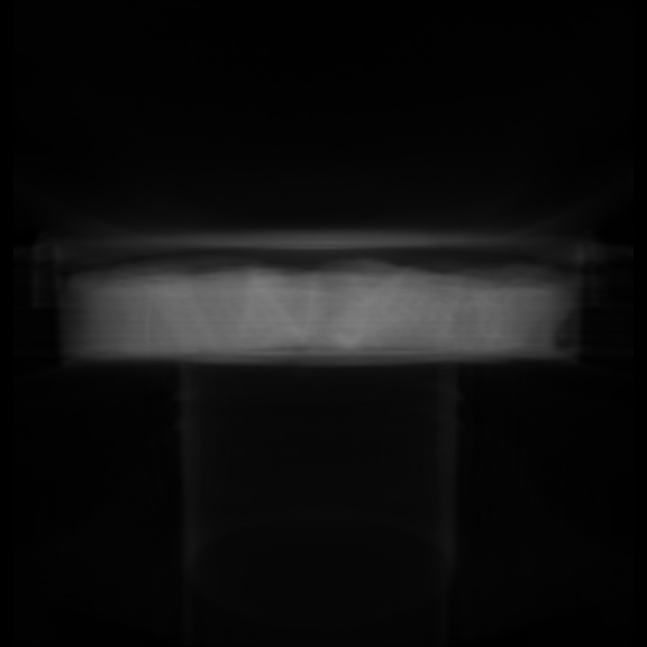}
    & \imgcell{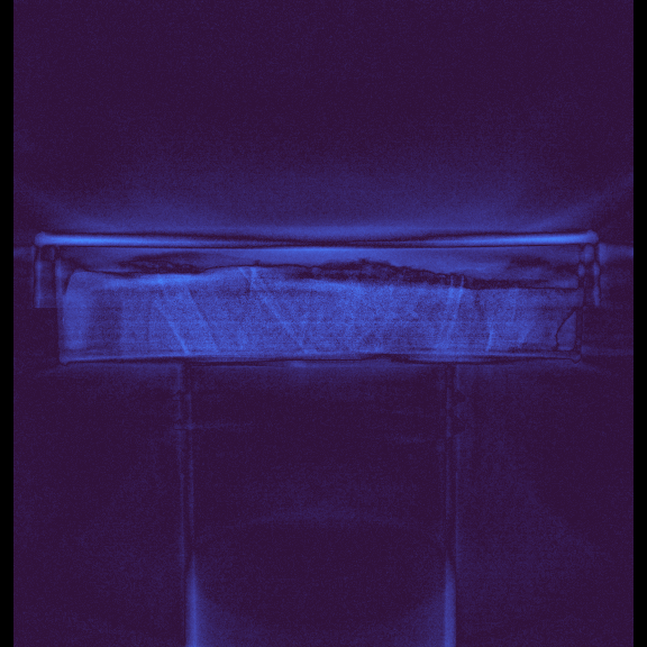}
    & \imgcell{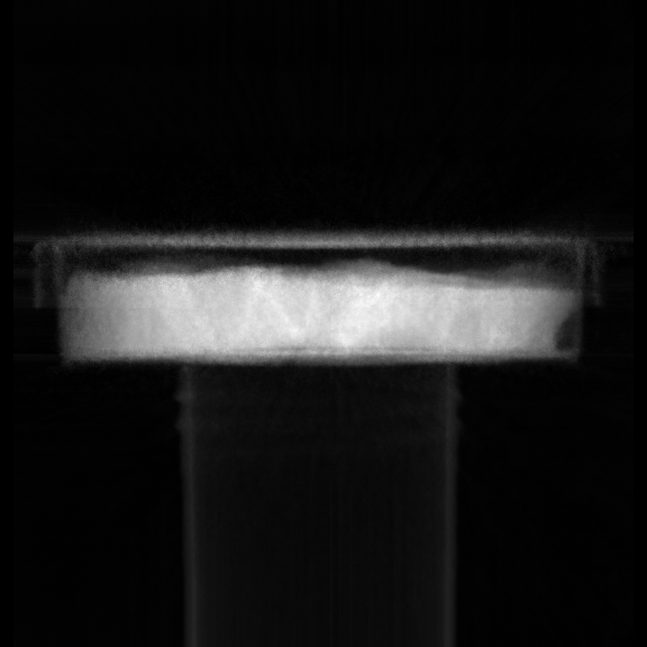}
    & \imgcell{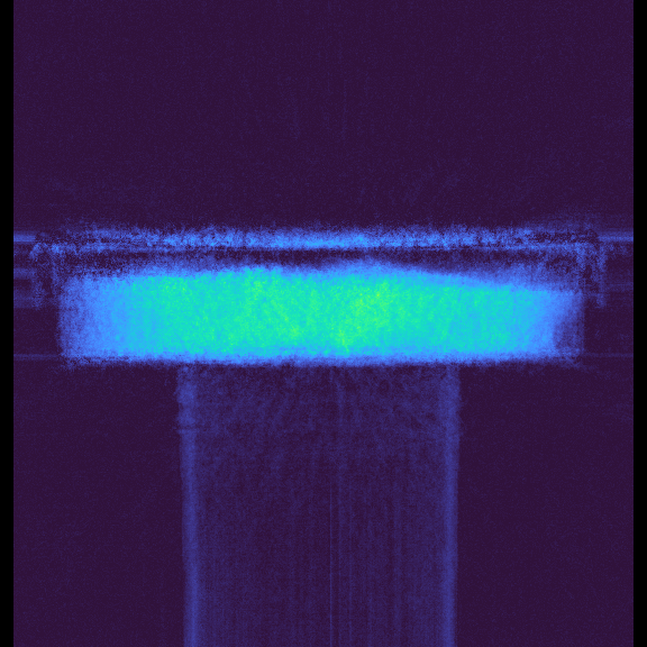}
    & \imgcell{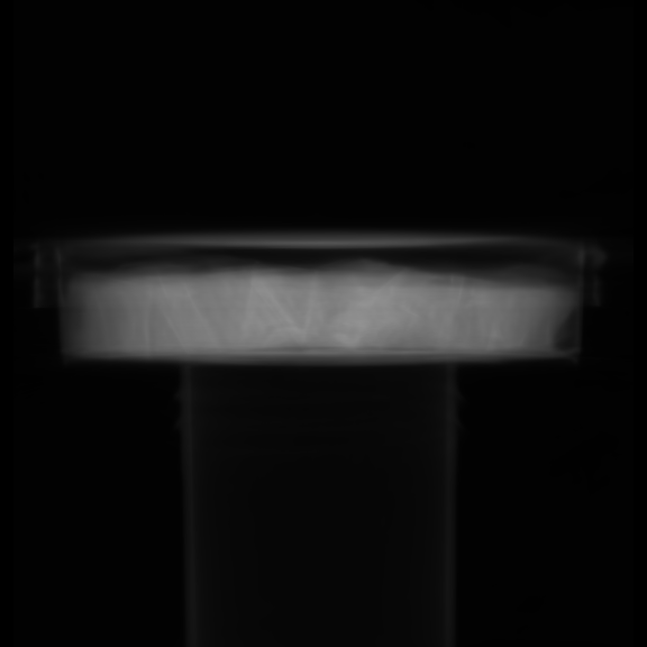}
    & \imgcell{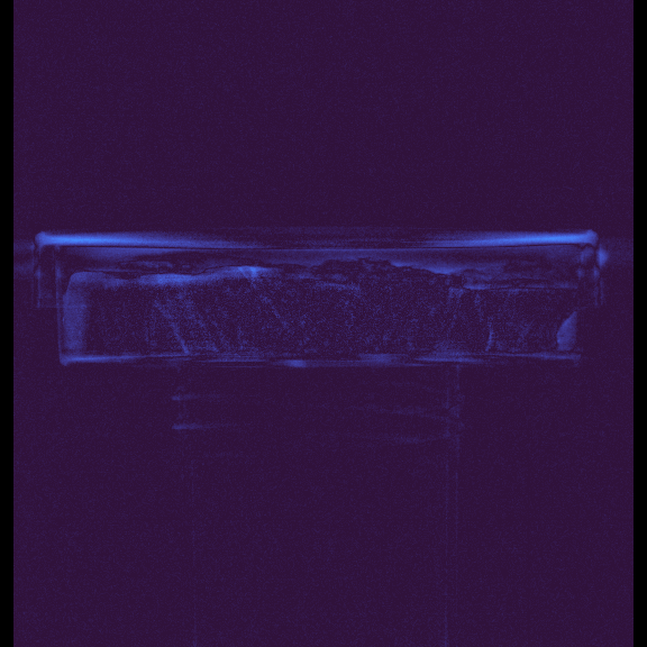}
    & \imgcell{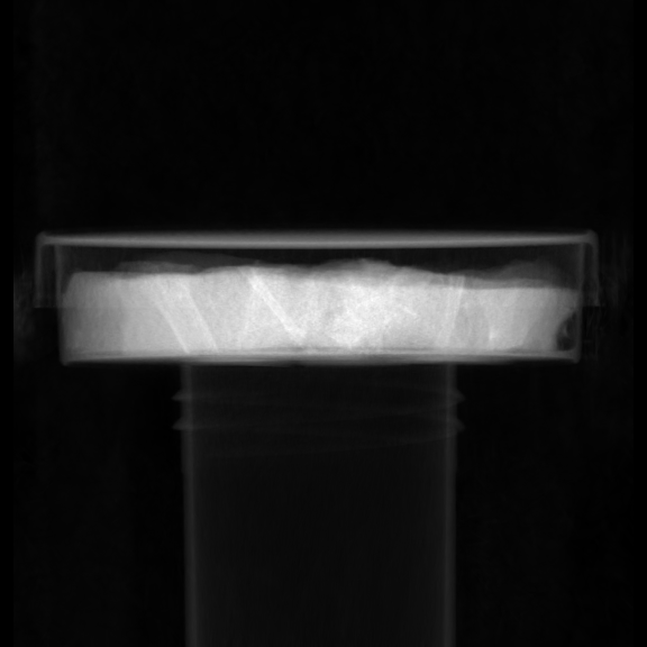}
    & \imgcell{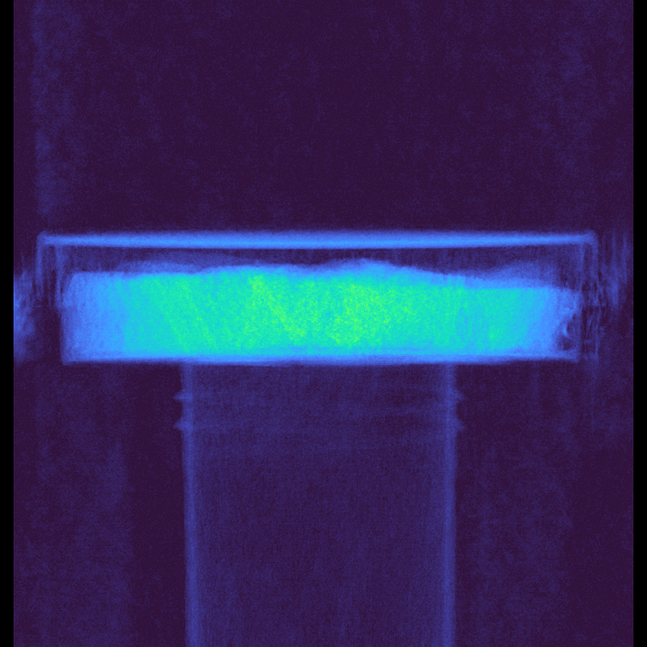}
    & \imgcell{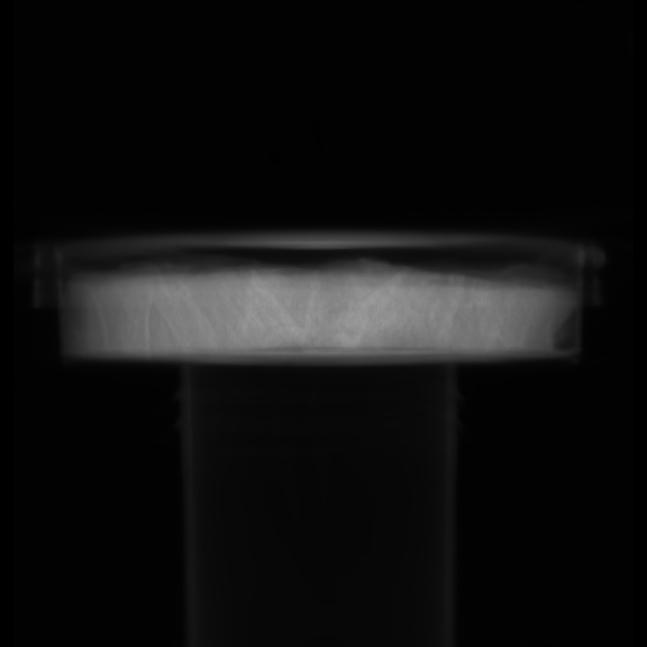}
    & \imgcell{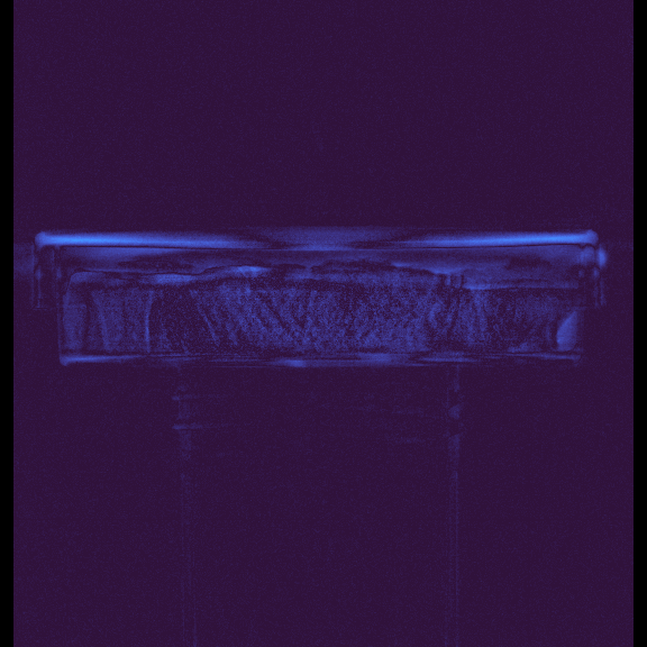}
    & \imgcell{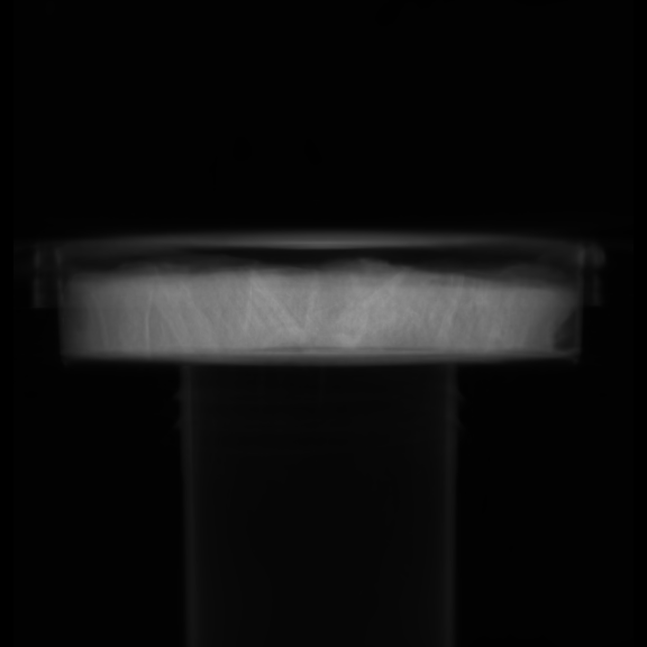}
    & \imgcell{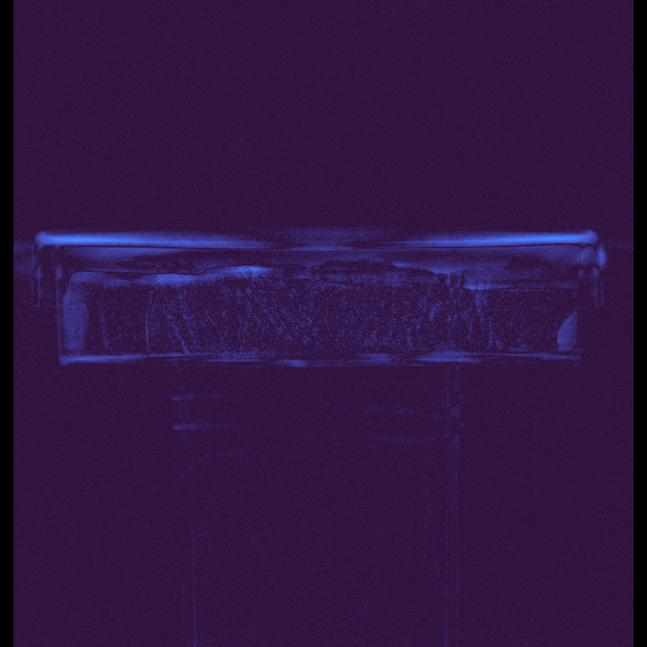}
    \\

    \enercell{120\,keV}
    & \imgcell{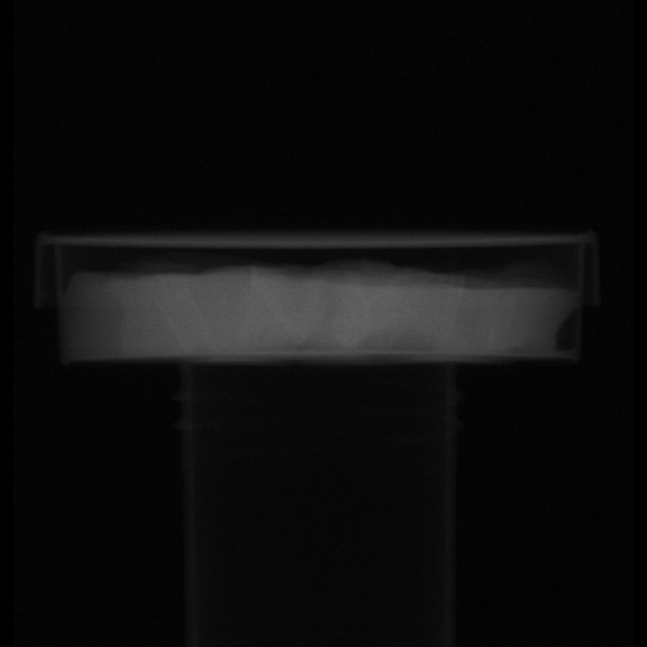}
    & \imgcell{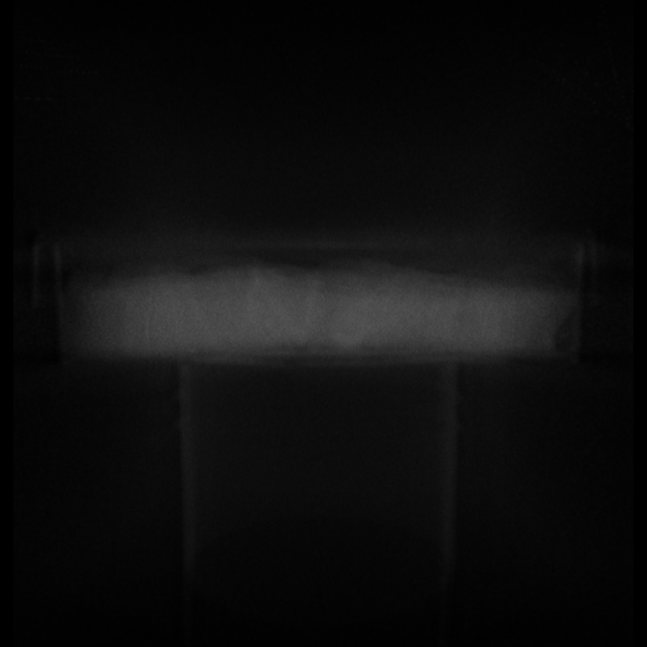}
    & \imgcell{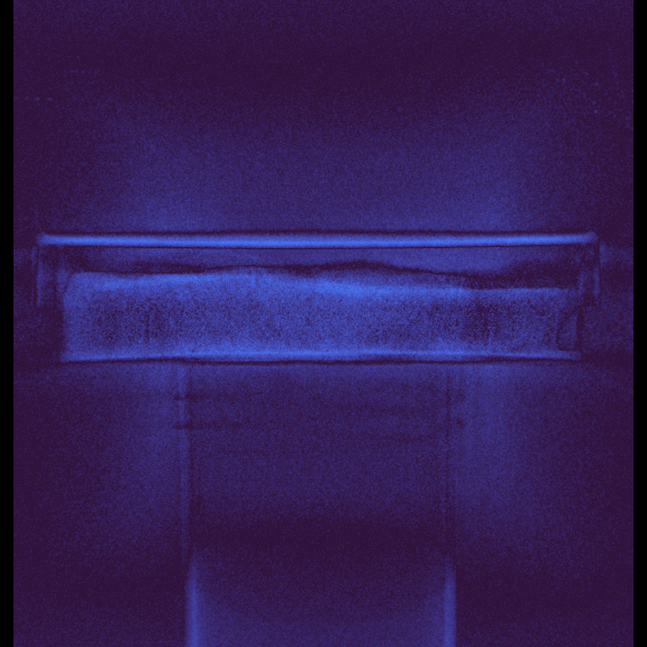}
    & \imgcell{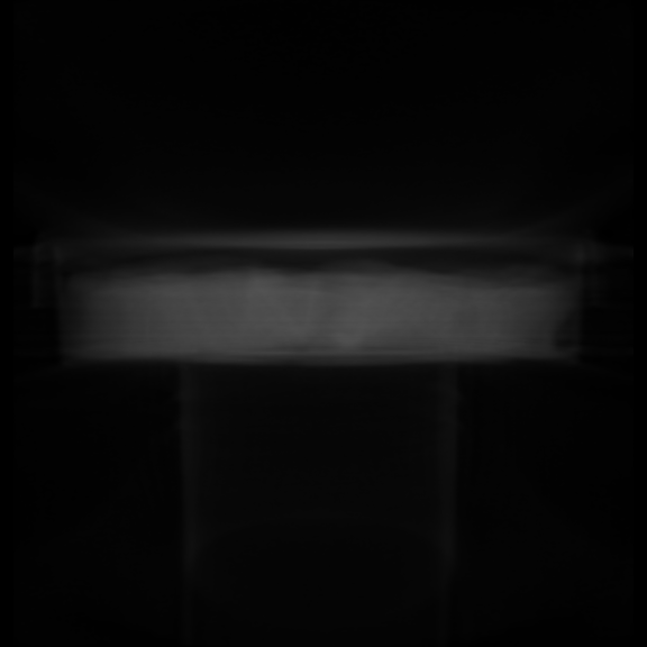}
    & \imgcell{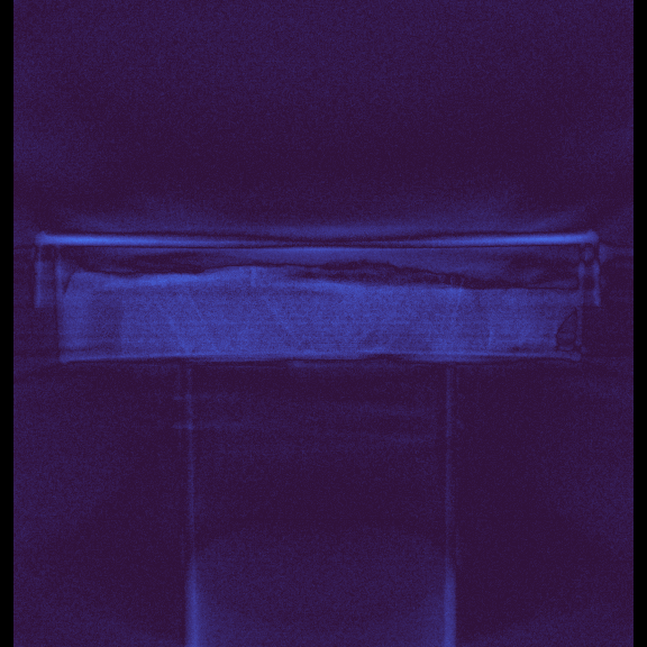}
    & \imgcell{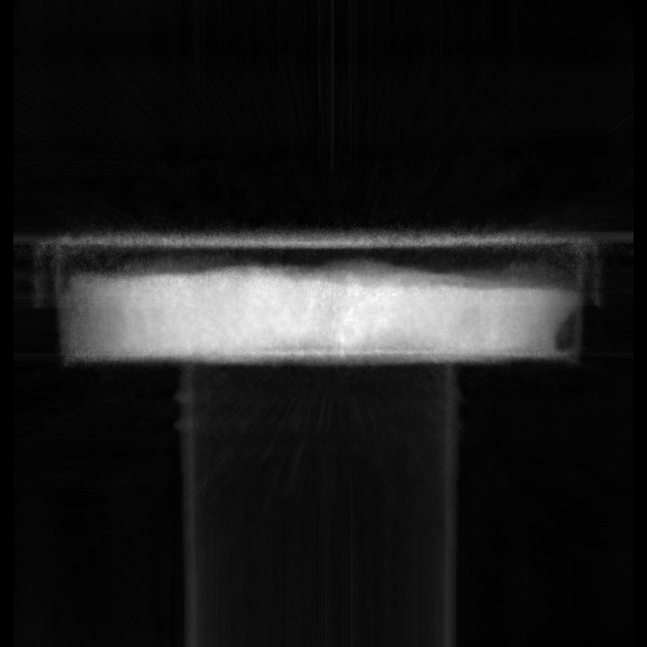}
    & \imgcell{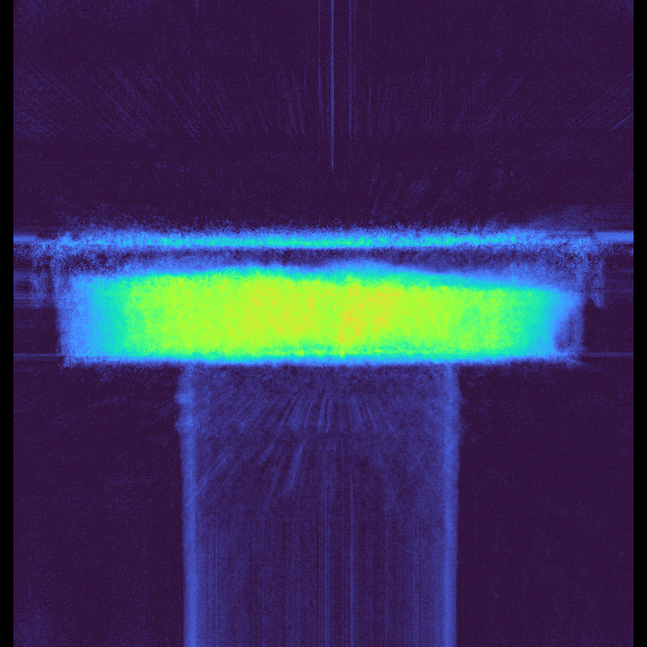}
    & \imgcell{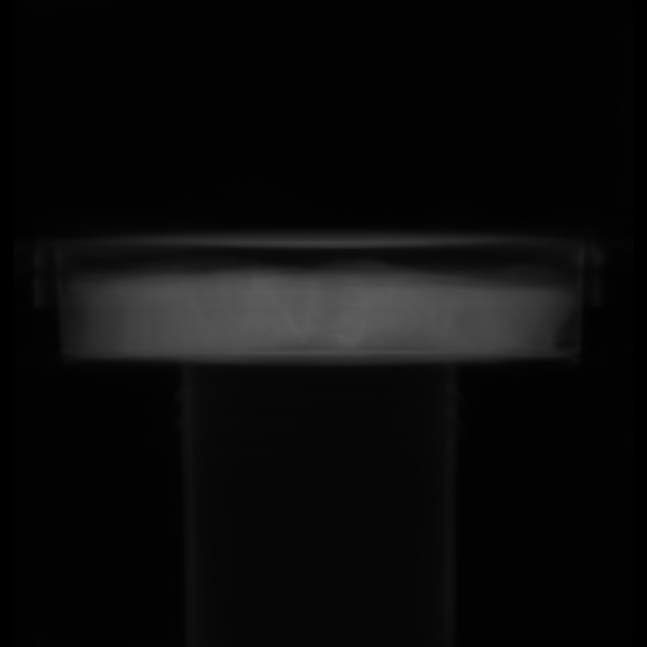}
    & \imgcell{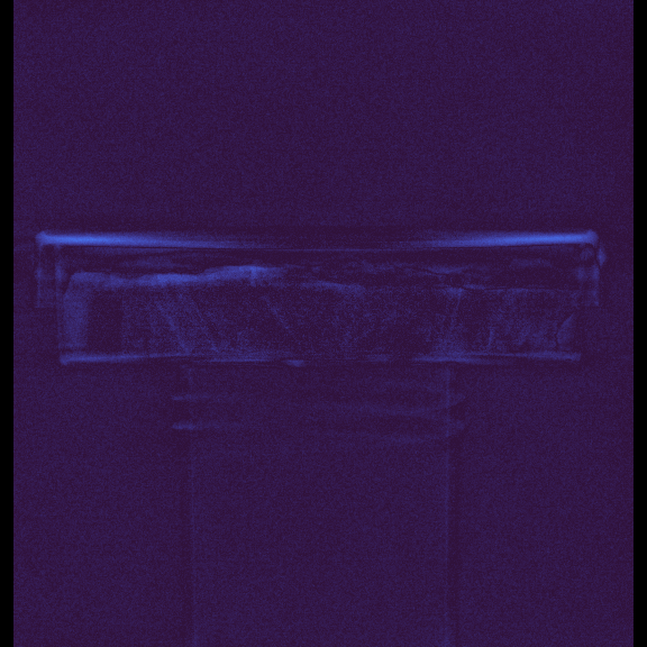}
    & \imgcell{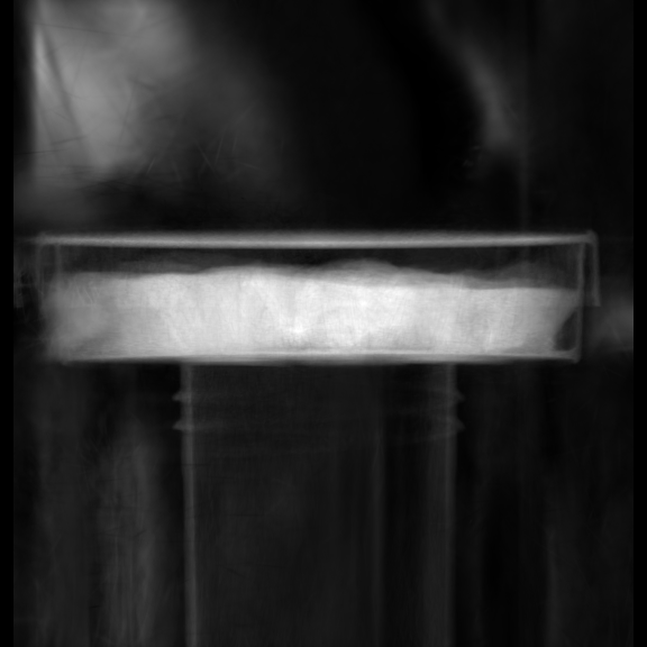}
    & \imgcell{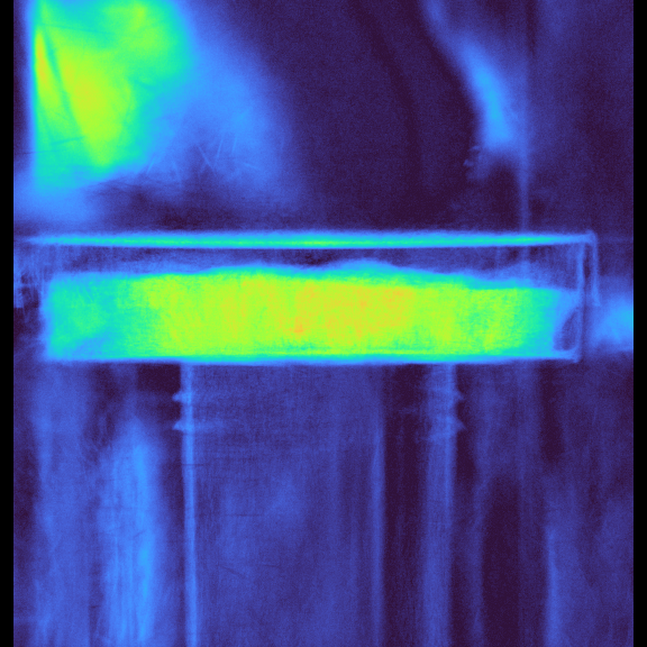}
    & \imgcell{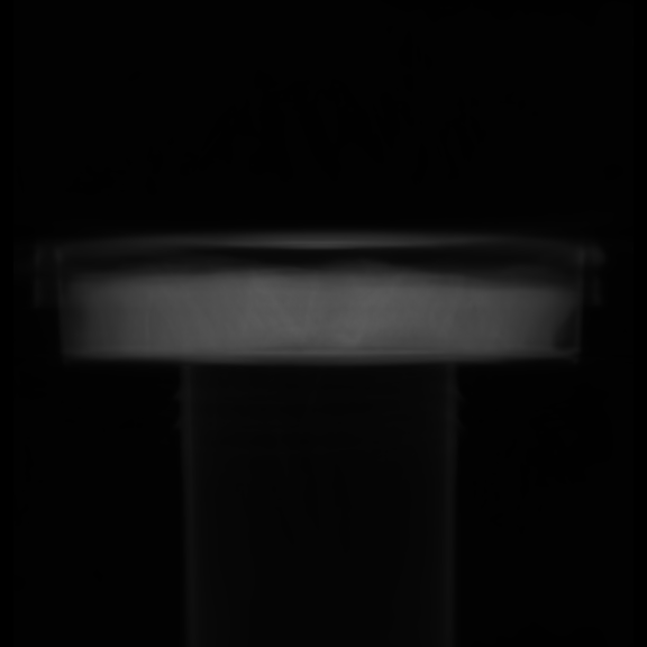}
    & \imgcell{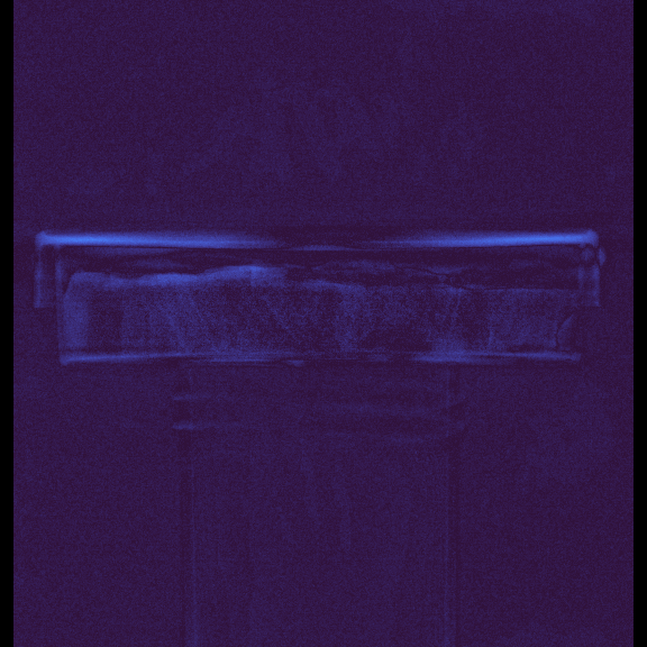}
    & \imgcell{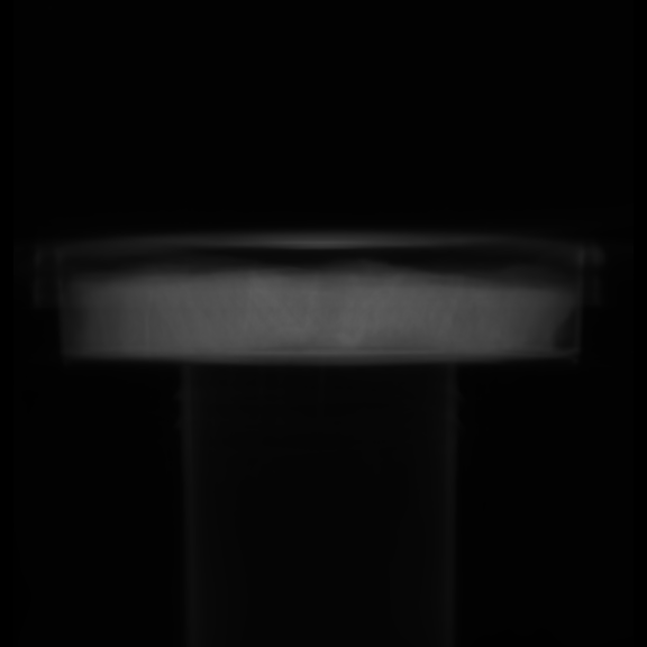}
    & \imgcell{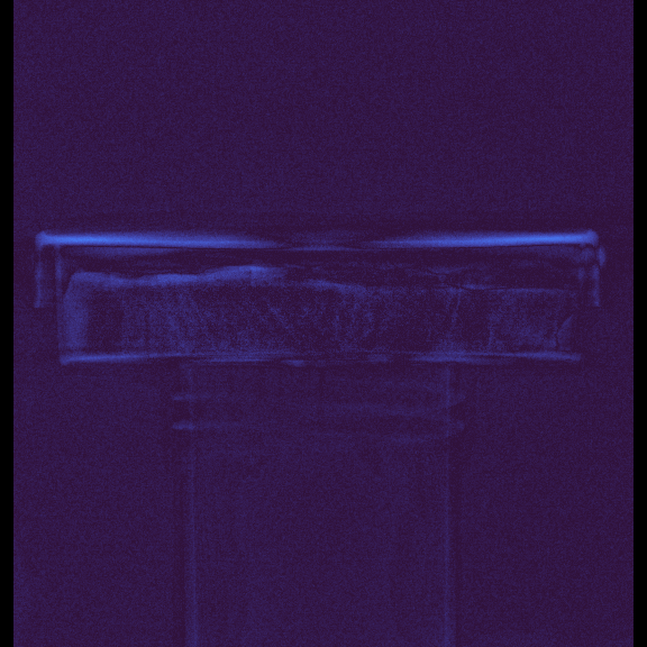}
    \\
    \addlinespace[3pt]
    & \multicolumn{15}{c}{\scriptsize Abs.\ error:\quad\raisebox{-0.35ex}{\errorcolorbarhoriz[3.2cm]}} \\
\end{tabular}
}
\end{figure*}
\section{Implementation details}
\label{sec:implementation_details}
We implemented our method in PyTorch with custom CUDA rasterization and voxelization kernels and trained on a
single Nvidia RTX 3090 for $30\mathrm{k}$ Adam iterations, with the spectral forward model evaluating the
energy integral using $64$ midpoint samples between $20$ and $140$~keV.
The point cloud is initialized with $M=50\mathrm{k}$ Gaussians sampled from the FDK volume and grows through
adaptive density control up to at most $500\mathrm{k}$ Gaussians.
We set $\lambda_{\mathrm{DSSIM}}=0.25$, $\lambda_{\mathrm{3DTV}}=0.005$ and
$\lambda_{\mathrm{TV},\mathrm{PE}}=\lambda_{\mathrm{TV},\mathrm{C}}=\lambda_{\mathrm{TV},\mathrm{K}}=0.05$,
and freeze the K-edge basis parameters for the first $2000$ iterations before optimizing them with a separate
Adam optimizer. 
\section{Evaluation}
\label{sec:evaluation}
We evaluate our method quantitatively on novel view synthesis, spectral CT reconstruction and basis
material decomposition, and qualitatively on projections, reconstructed slices and material maps.
In addition, we run ablations to show the effect of the individual components of our model.
\subsection{Comparison with baseline methods}
\label{sec:baselines}
To show the potential of our approach we compare it against both traditional and learning-based
methods. As traditional baselines we use FDK~\cite{Feldkamp:84} and SART~\cite{ANDERSEN198481} channel-wise,
together with the iterative solvers CGLS~\cite{Soleimani2015IterativeCT} and
PDHG-TV~\cite{Chambolle2011}, for which we rely on the implementations in the Core Imaging Library
(CIL)~\cite{pasca_2026_20733546, 10.1098/rsta.2020.0192}.
These methods only reconstruct volumes, so for novel view synthesis we first reconstruct the volume
and then render the projections with TIGRE~\cite{Biguri_2016}. To the best of our knowledge, no publicly available implementation of a radiance-based representation for spectral CT reconstruction currently exists.
We therefore select learning-based models from the closest related domain, conventional
single-channel CT reconstruction, and train a separate model for every energy channel of a dataset.
From the neural field category we use IntraTomo~\cite{9710599} and
SAX-NeRF~\cite{cai2024structureawaresparseviewxray3d}, and from the 3DGS category
X-Gaussian~\cite{cai2024radiativegaussiansplattingefficient}, R2-Gaussian~\cite{r2_gaussian} and
X-Field~\cite{wang2025xfieldphysicallygroundedrepresentation}.
X-Gaussian and X-Field are designed only for novel view synthesis and cannot produce volumes, so
they are not included in the CT reconstruction comparison.
All results are obtained with the publicly available implementations provided by the respective
authors. Our own method is reported in two variants.
\textit{Ours (2)} uses two basis materials, the photoelectric and the Compton basis, which describe
the smooth energy dependence of the attenuation of common materials.
\textit{Ours (3)} adds a third learnable K-edge basis for objects that contain contrast agents or
heavy elements with a K-edge inside the scanned energy range, where the two smooth bases cannot
describe the sudden jump in attenuation.
\subsection{Quantitative results}
\label{sec:quantitative_results}
We now compare our method quantitatively against the traditional and learning-based baselines on all
three tasks, namely novel view synthesis, spectral CT reconstruction and basis material decomposition.
\noindent\textbf{Novel view synthesis}
The upper block of Table~\ref{tab:quail-metrics} reports PSNR, SSIM and LPIPS for novel view
synthesis on the $100$ held-out test projections of the bird chest
dataset~\cite{BIRDCHEST_meaney_2024_11388131}, so it shows how well each method predicts projection
images from viewing angles that were not used during training. Our approach outperforms recent learning-based approaches, with the exception
of R$^2$-Gaussian~\cite{r2_gaussian}, which matches our accuracy but at the
cost of one model per energy channel and hence three times as many Gaussians in total across all channels.
In comparison to the traditional methods we reach better PSNR and SSIM values on all datasets. When compared to the traditional methods, we reach better PSNR and SSIM values on all datasets,
and for LPIPS we are better than every other method except FDK, which still reaches the best value. The results shows that our single spectral representation predicts novel views as well as the best
per-channel methods, while describing all energy channels at once.
\begin{figure*}[htbp!]
\centering
\caption{Spectral CT reconstruction qualitative results on the multi-energy bird chest dataset~\cite{BIRDCHEST_meaney_2024_11388131}. Columns show the compared methods, rows the axial slice and an outside view at each energy.}
\label{fig:volume-quail-images}
\setlength{\tabcolsep}{1pt}
\renewcommand{\arraystretch}{0.9}
\renewcommand{\imgcell}[1]{\includegraphics[width=0.098\textwidth]{#1}}
\renewcommand{\methodcell}[1]{{\scriptsize\bfseries #1}}
\newcommand{\rowlabel}[2]{\raisebox{0.22cm}[0pt][0pt]{\scriptsize\bfseries\shortstack[l]{#1\\#2}}}
\scalebox{1}{
\begin{adjustbox}{center,max totalsize={\textwidth}{\textheight}}
    \begin{tabular}{l@{\hspace{3pt}}cccccccccc}
        &
        \methodcell{GT} &
        \methodcell{FDK} &
        \methodcell{SART} &
        \methodcell{CGLS} &
        \methodcell{PDHG-TV} &
        \methodcell{IntraTomo} &
        \methodcell{SAX-NeRF} &
        \methodcell{R2-Gaussian} &
        \methodcell{Ours (2)} &
        \methodcell{Ours (3)} \\
        &
        &
        {\scriptsize\cite{Feldkamp:84}} &
        {\scriptsize\cite{ANDERSEN198481}} &
        {\scriptsize\cite{Soleimani2015IterativeCT}} &
        {\scriptsize\cite{Chambolle2011}} &
        {\scriptsize\cite{9710599}} &
        {\scriptsize\cite{cai2024structureawaresparseviewxray3d}} &
        {\scriptsize\cite{r2_gaussian}} &
        &
        \\
        \addlinespace[2pt]

        \rowlabel{50\,keV}{Axial}
        & \imgcell{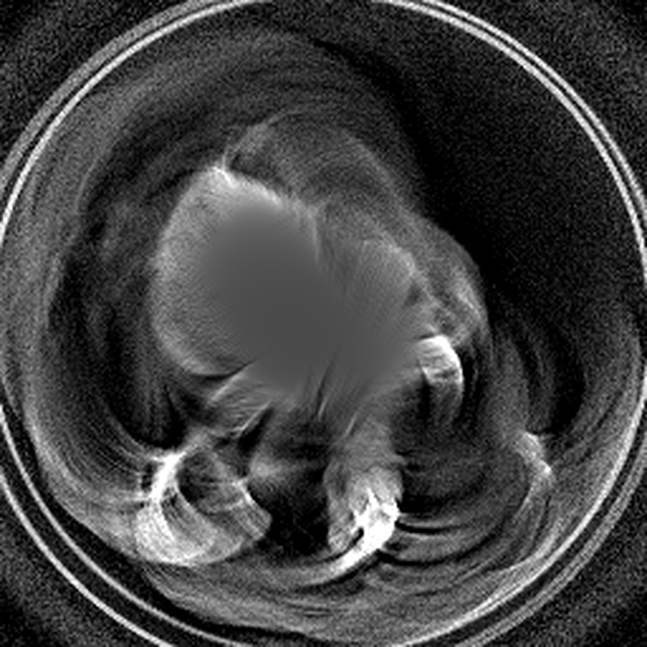}
        & \imgcell{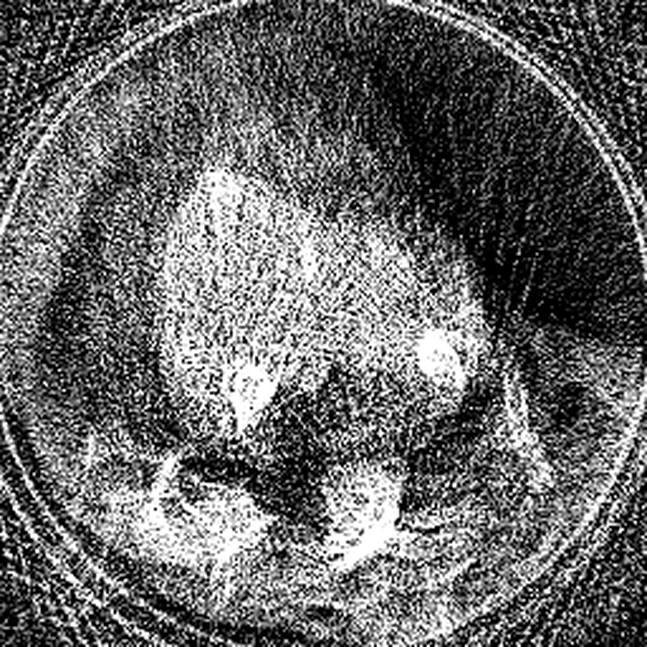}
        & \imgcell{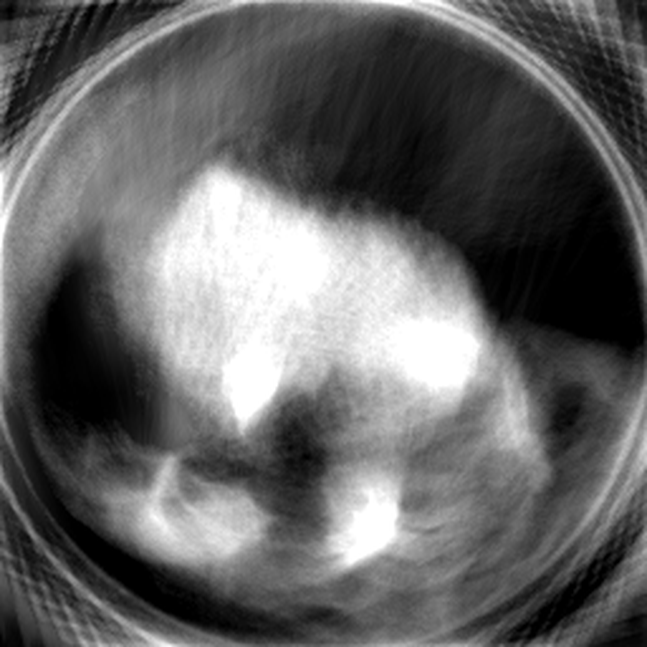}
        & \imgcell{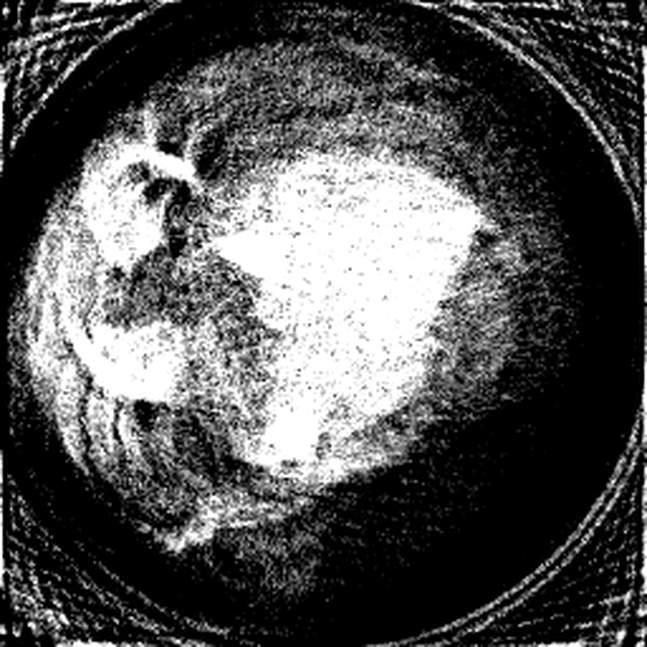}
        & \imgcell{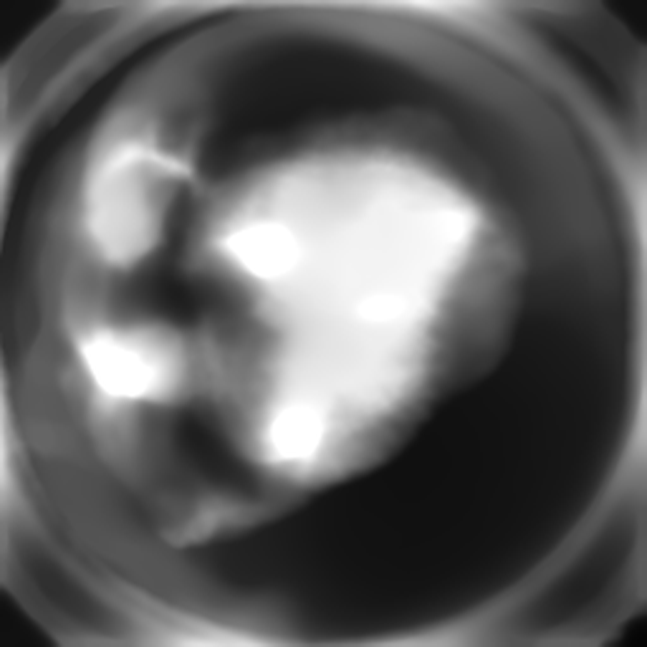}
        & \imgcell{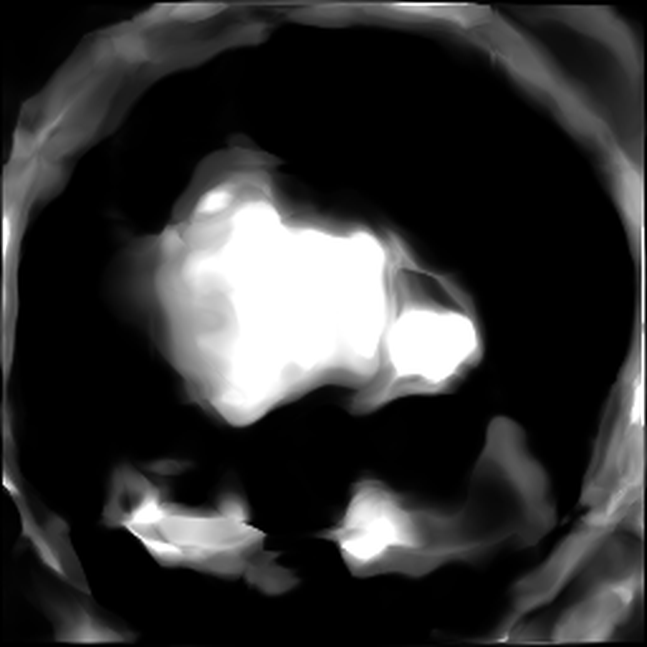}
        & \imgcell{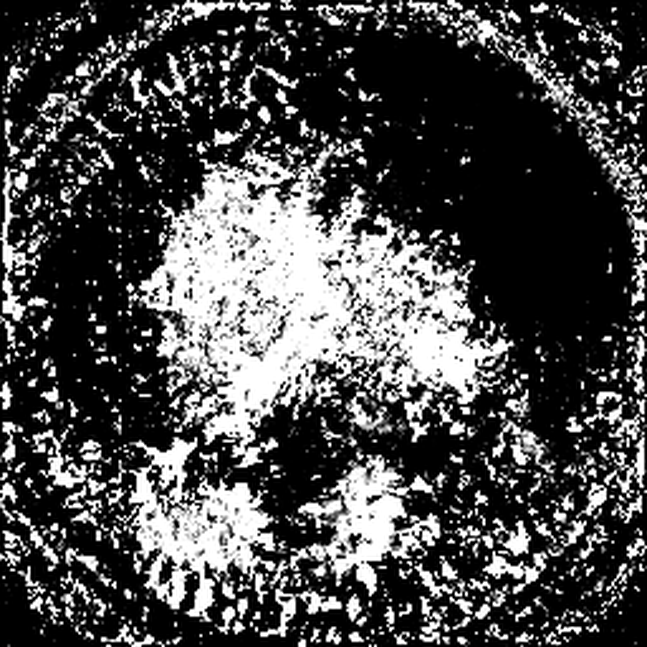}
        & \imgcell{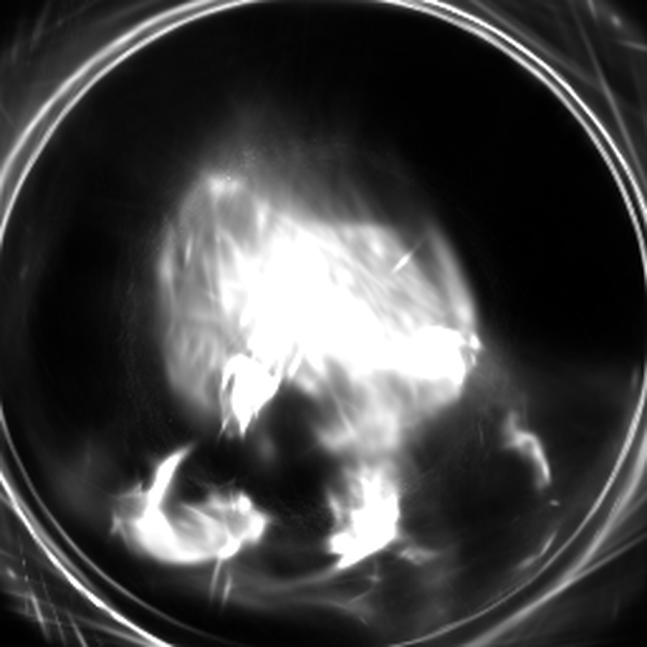}
        & \imgcell{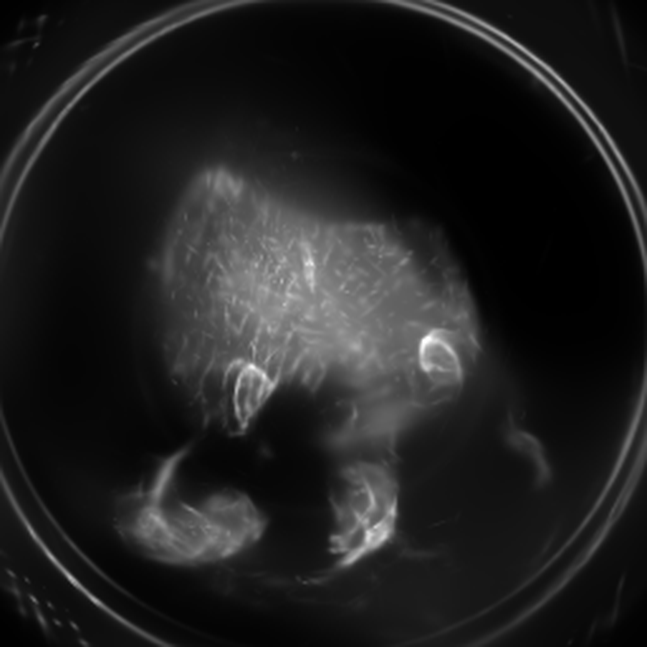}
        & \imgcell{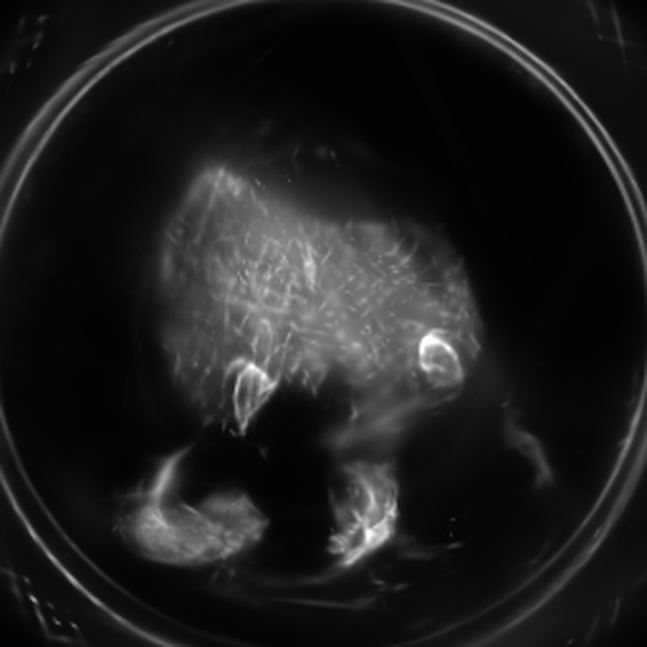}
        \\[-0.5mm]

        \rowlabel{50\,keV}{Outside}
        & \imgcell{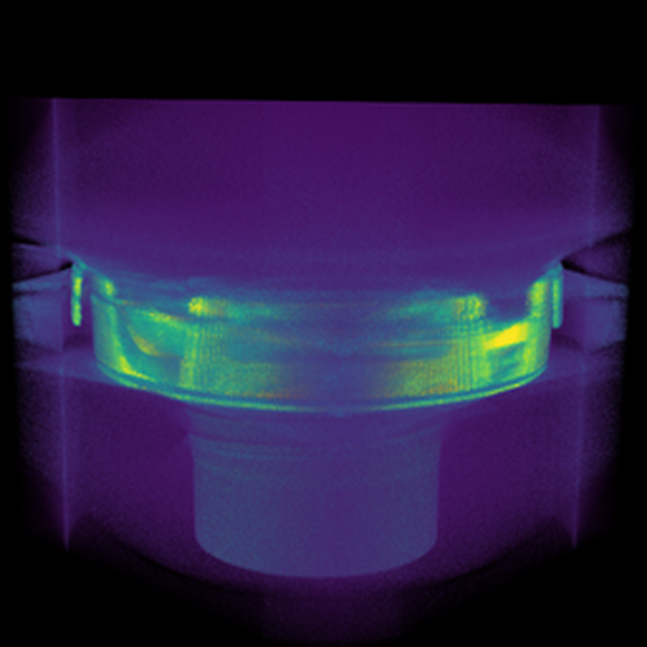}
        & \imgcell{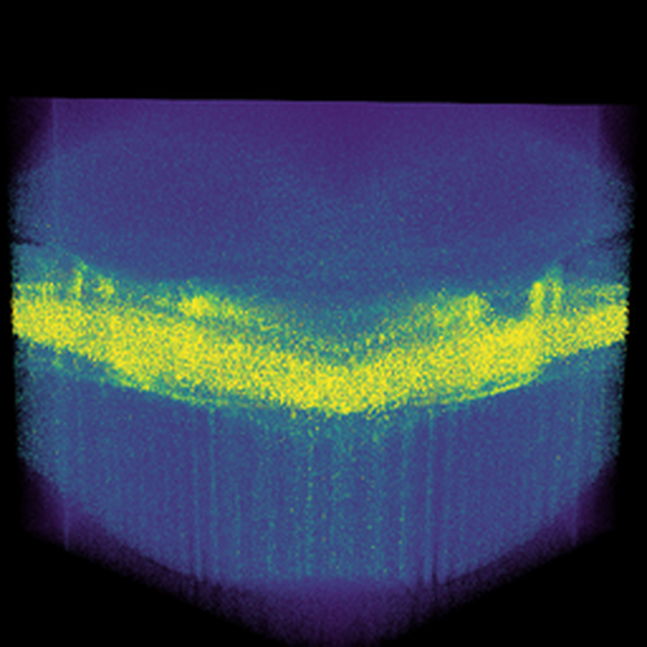}
        & \imgcell{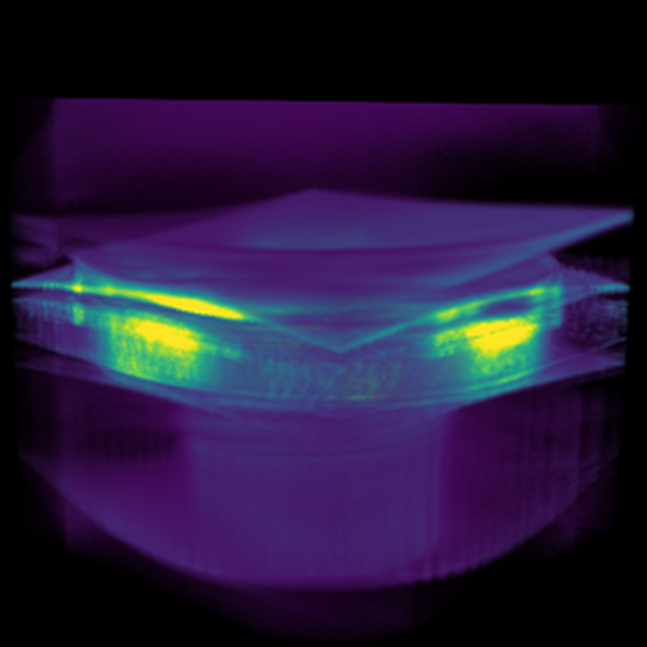}
        & \imgcell{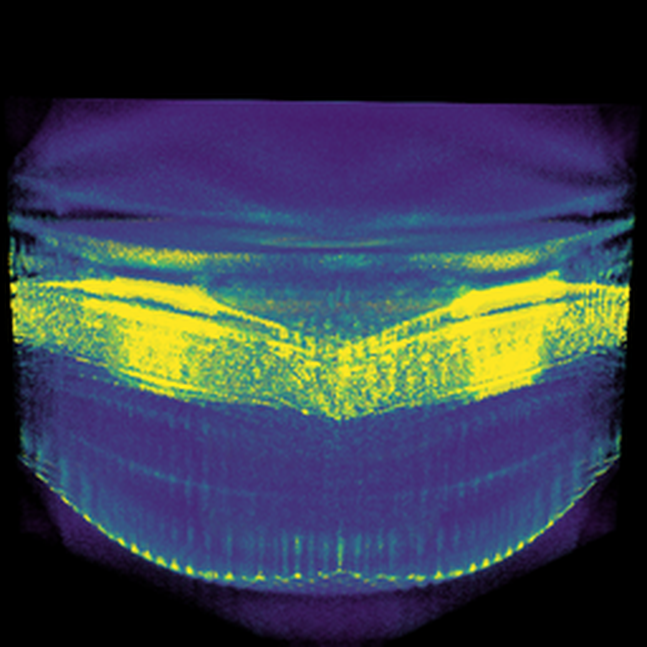}
        & \imgcell{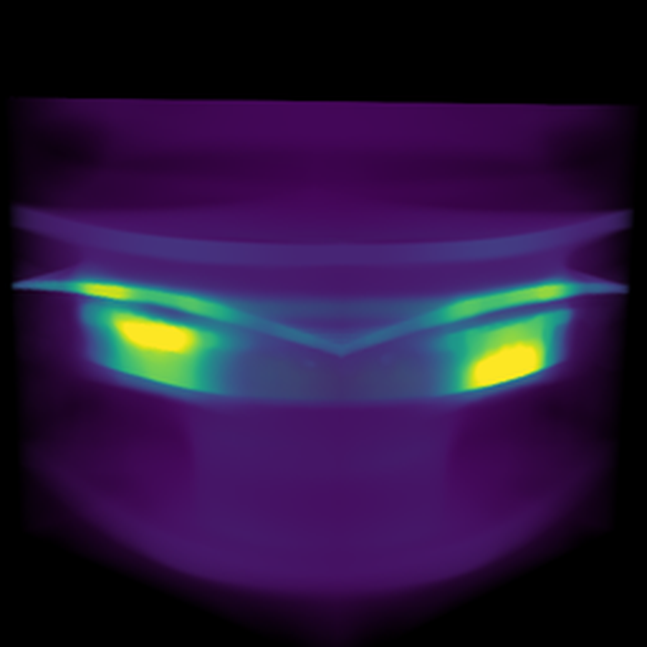}
        & \imgcell{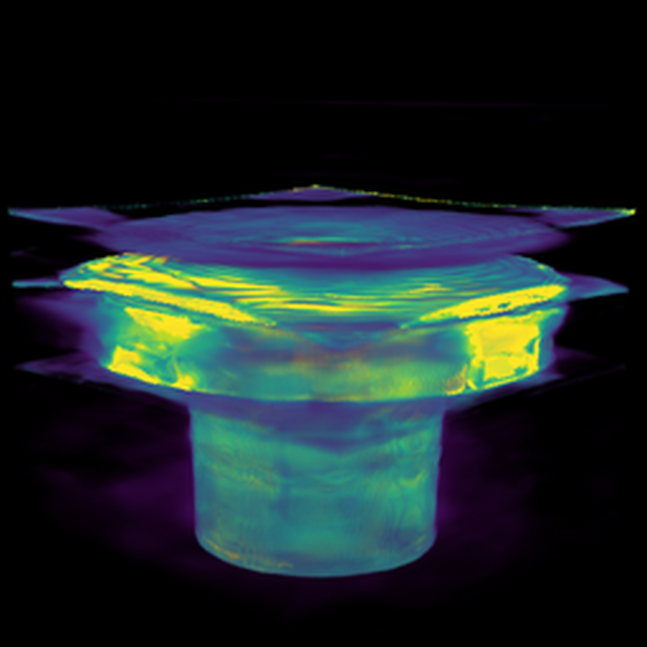}
        & \imgcell{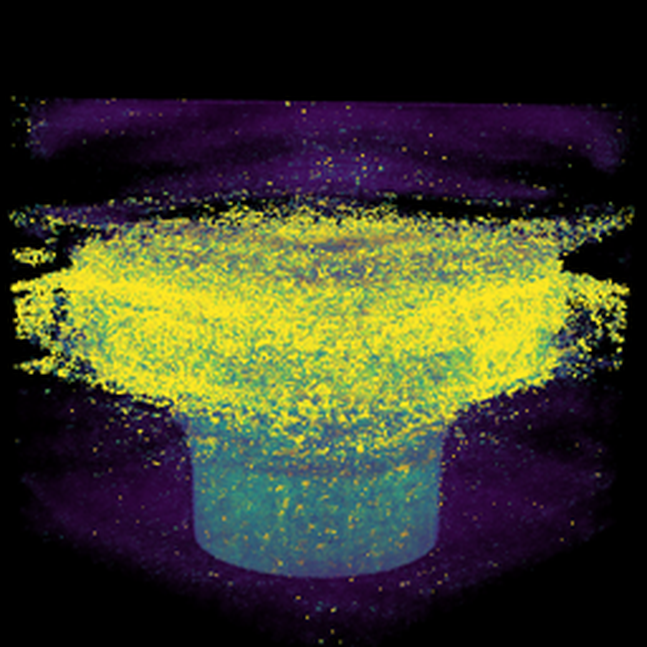}
        & \imgcell{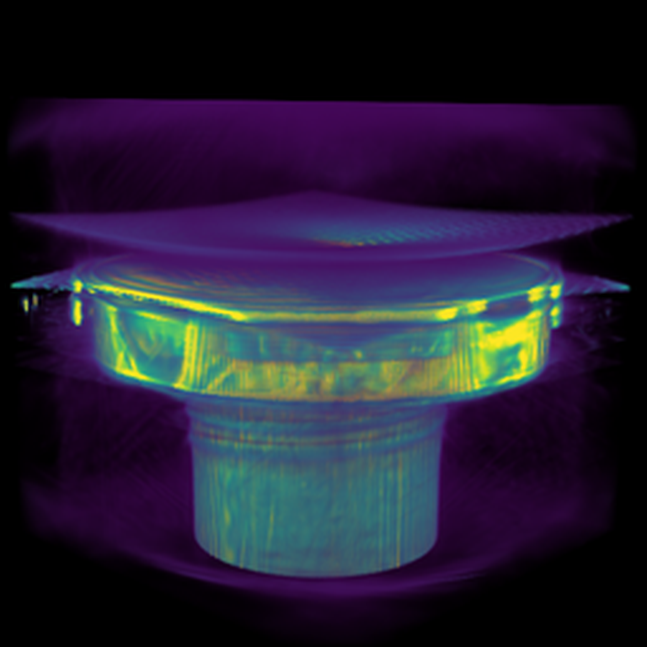}
        & \imgcell{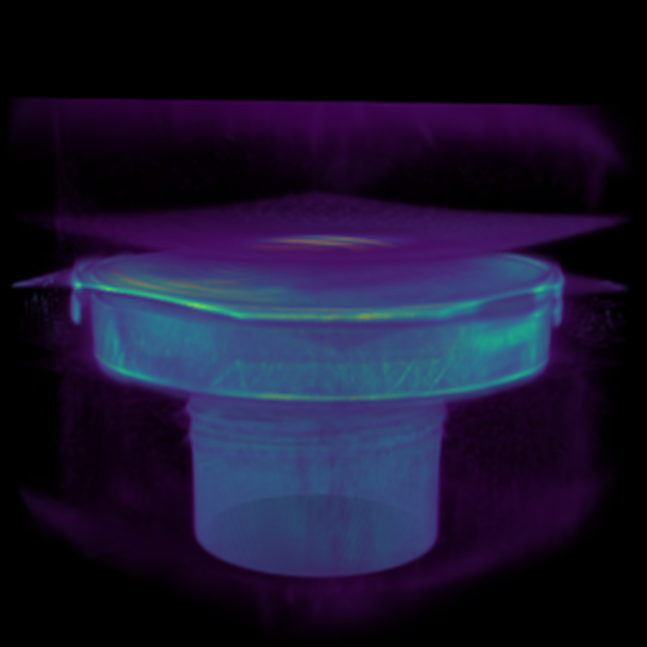}
        & \imgcell{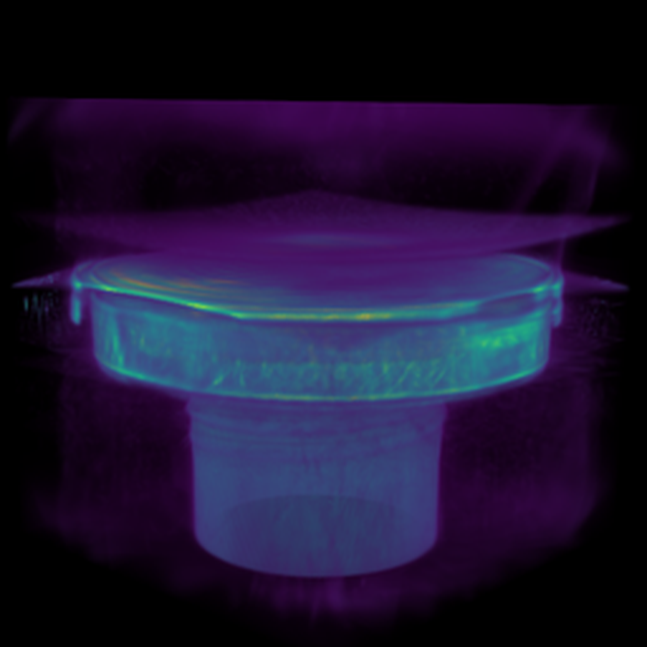}
        \\
        \addlinespace[3pt]

        \rowlabel{80\,keV}{Axial}
        & \imgcell{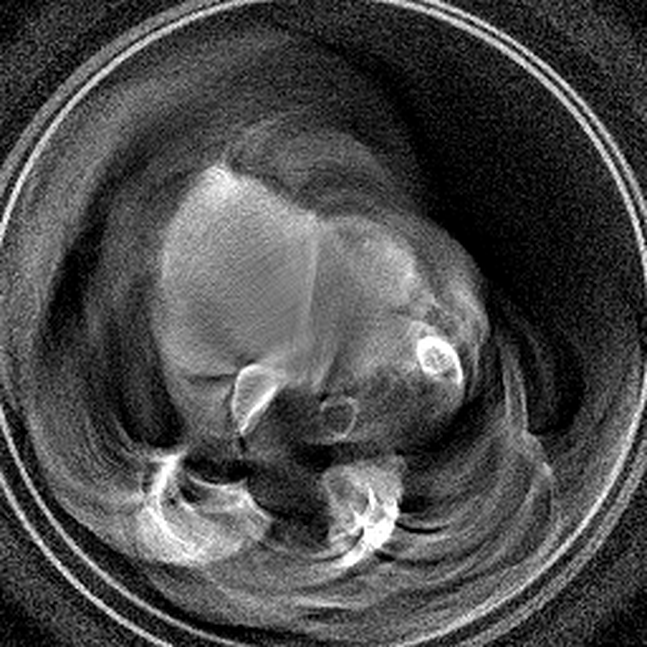}
        & \imgcell{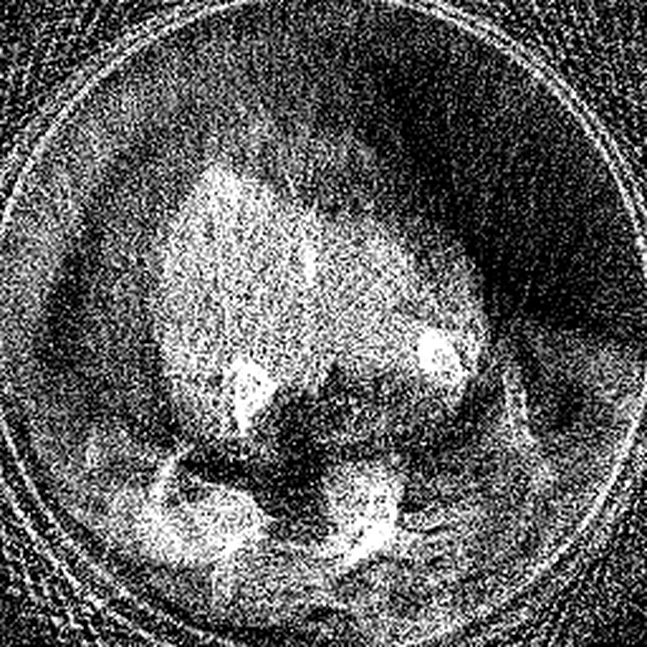}
        & \imgcell{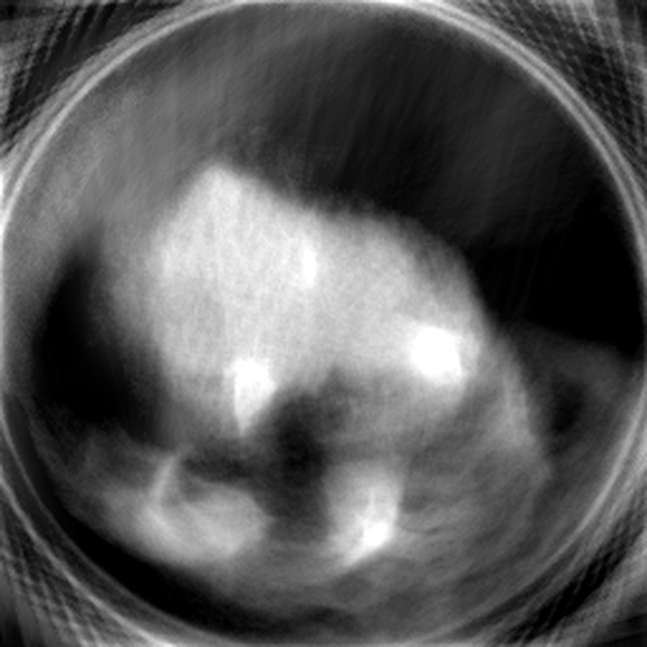}
        & \imgcell{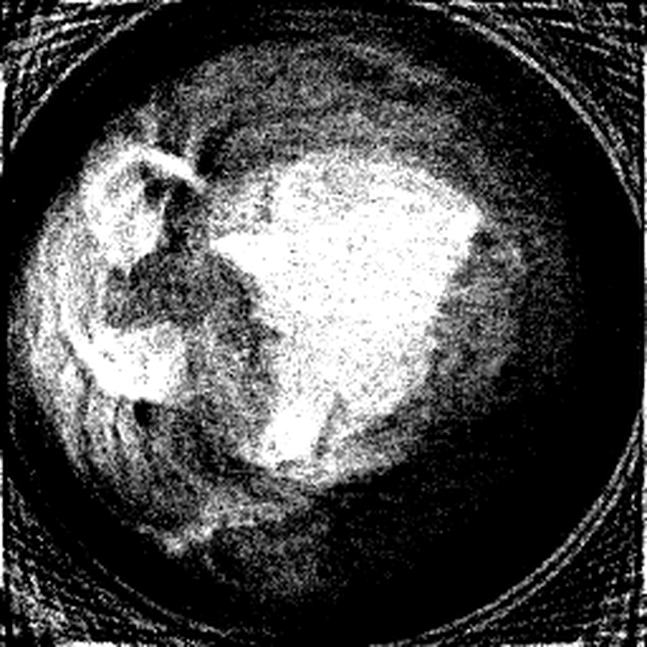}
        & \imgcell{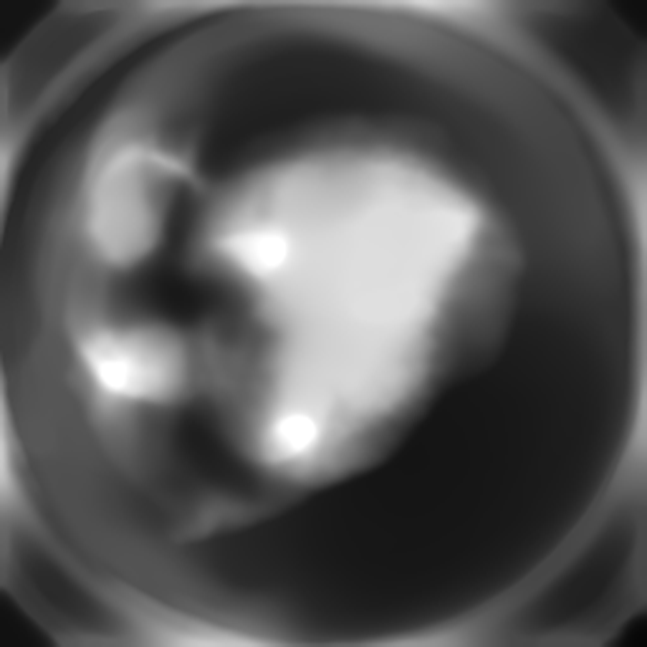}
        & \imgcell{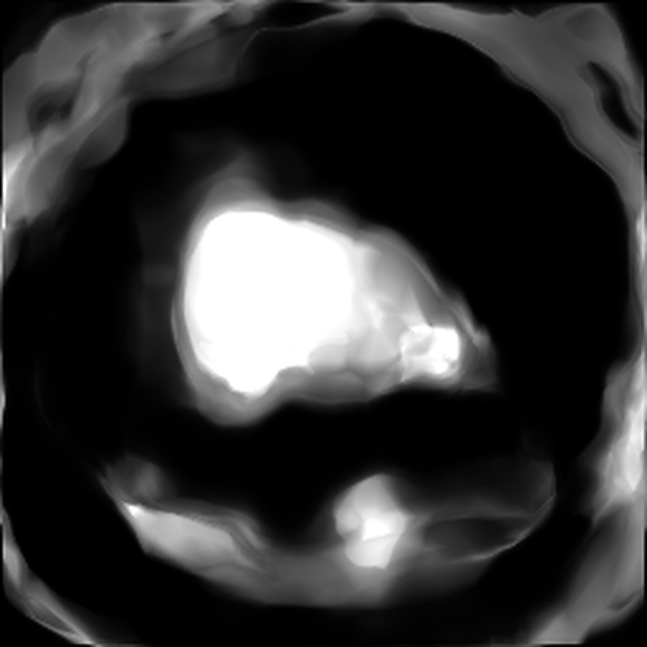}
        & \imgcell{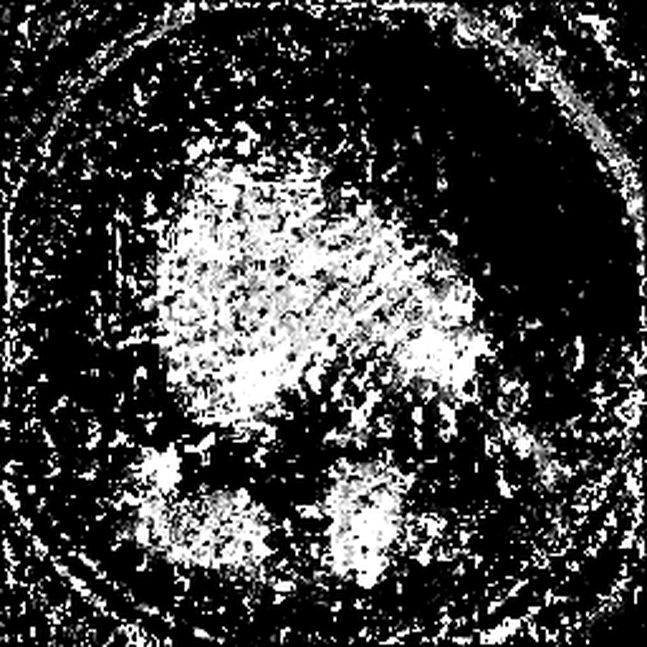}
        & \imgcell{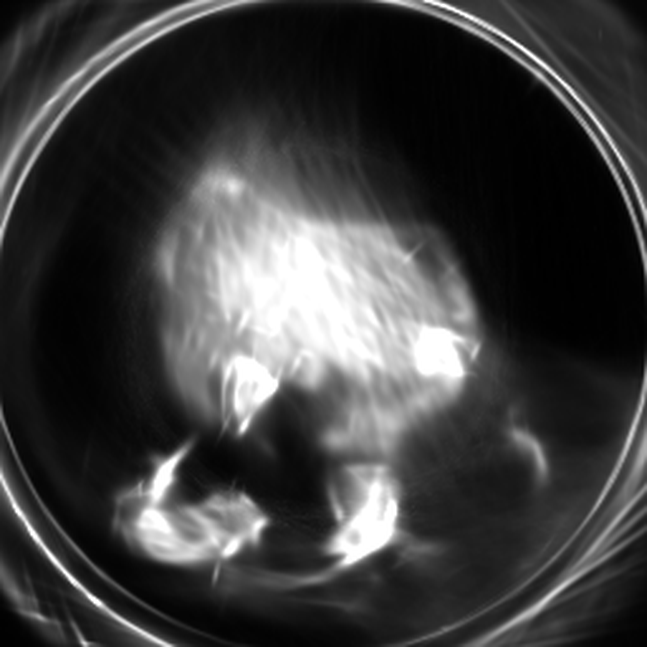}
        & \imgcell{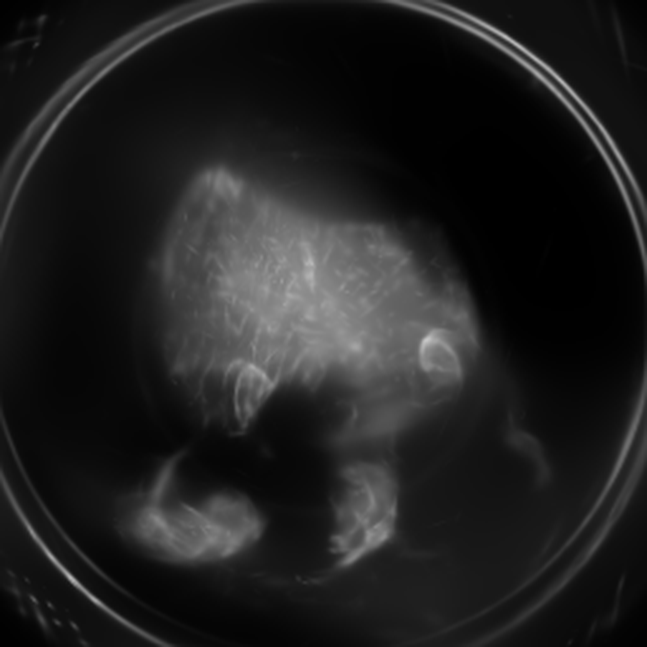}
        & \imgcell{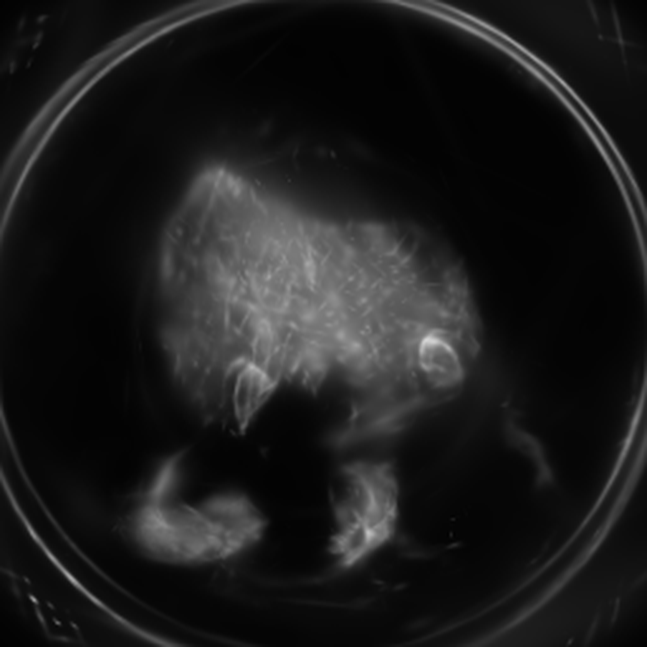}
        \\[-0.5mm]

        \rowlabel{80\,keV}{Outside}
        & \imgcell{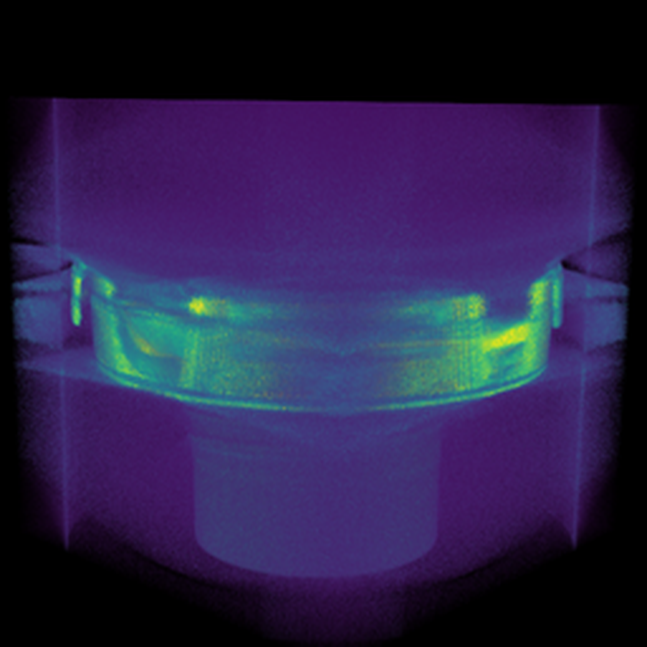}
        & \imgcell{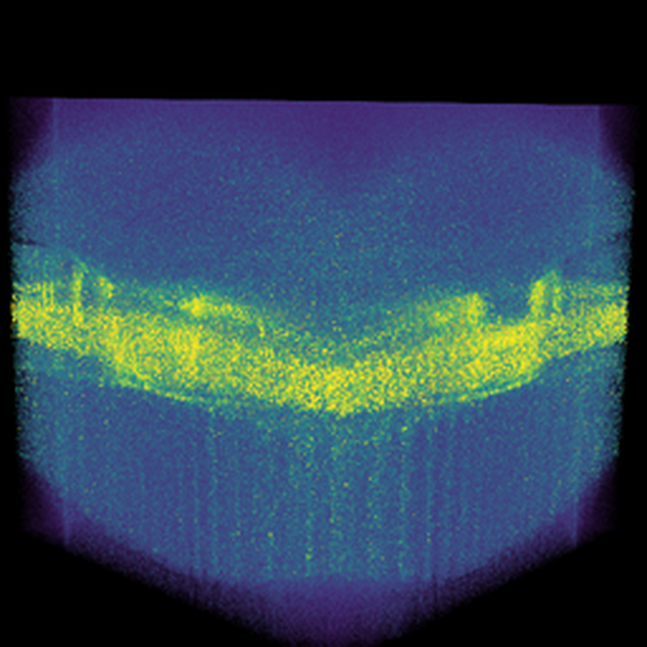}
        & \imgcell{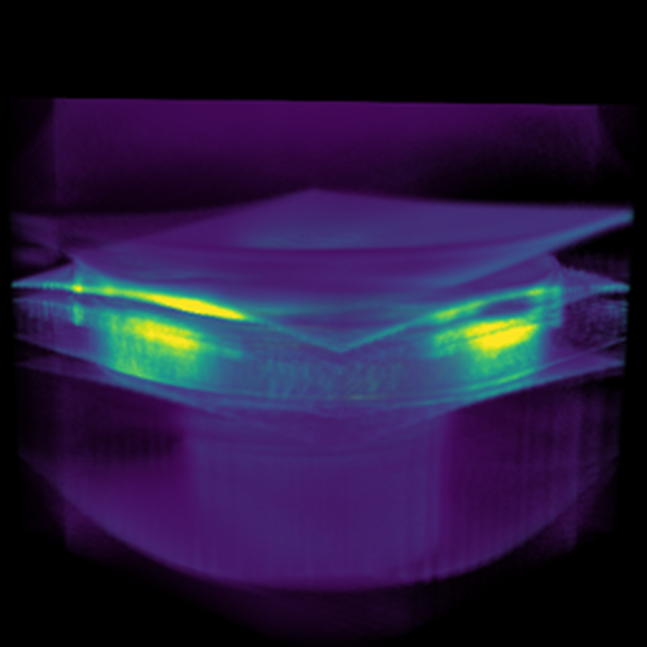}
        & \imgcell{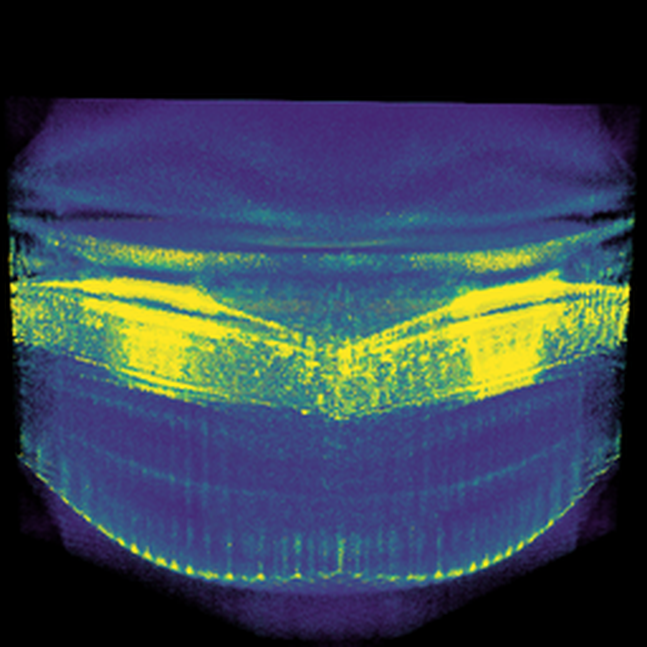}
        & \imgcell{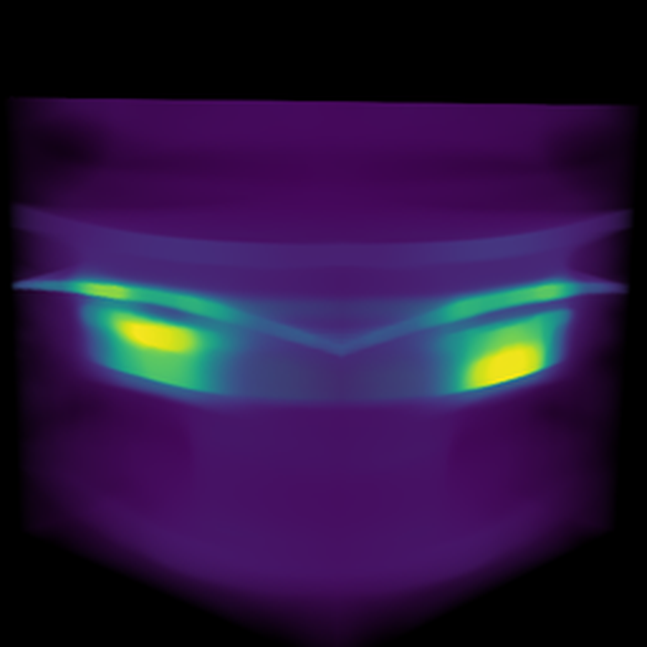}
        & \imgcell{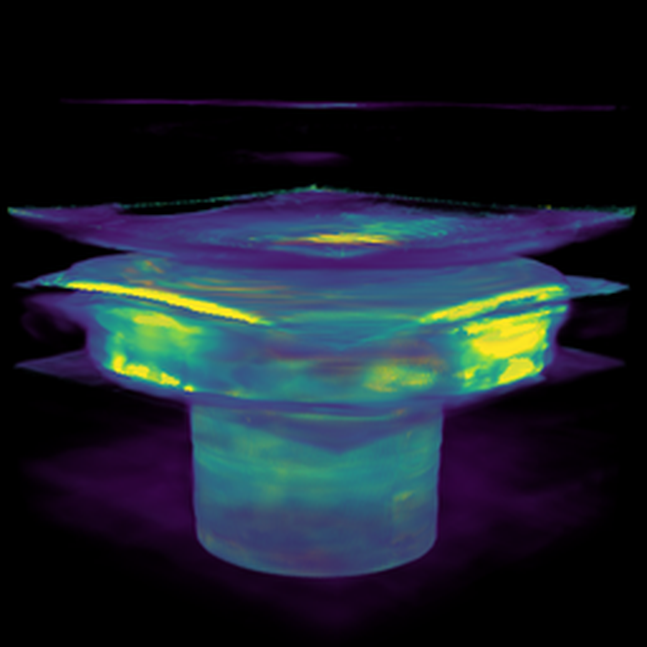}
        & \imgcell{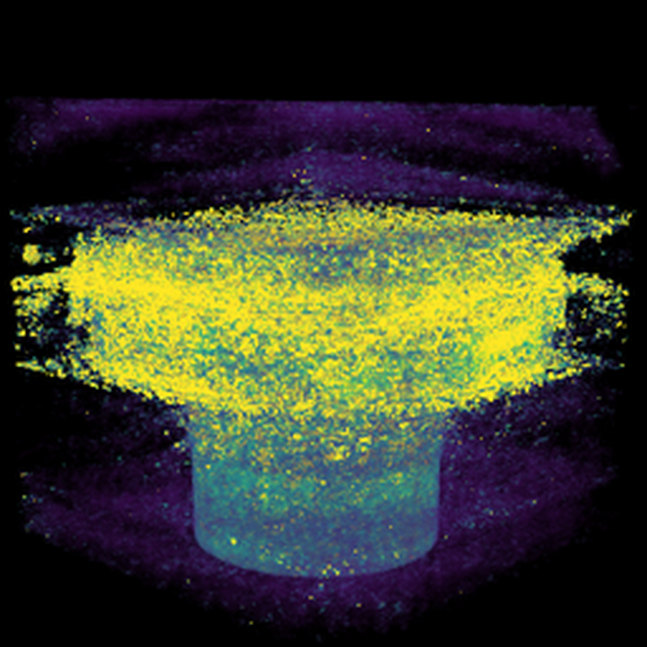}
        & \imgcell{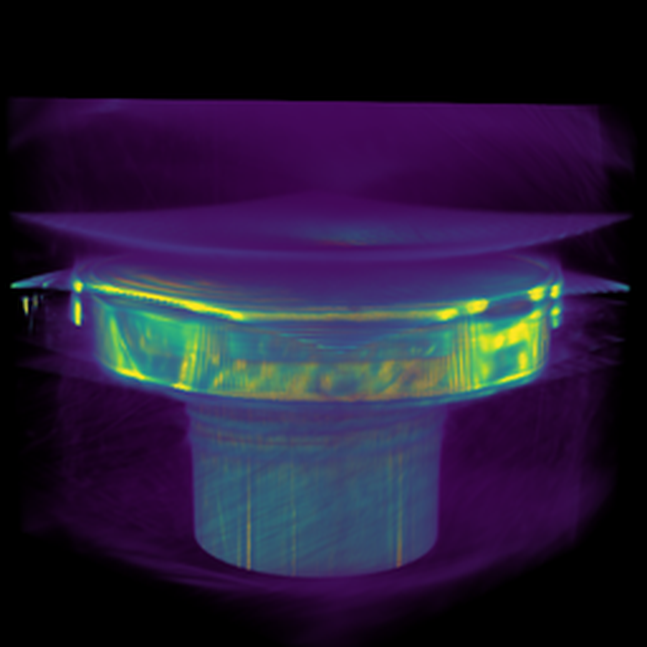}
        & \imgcell{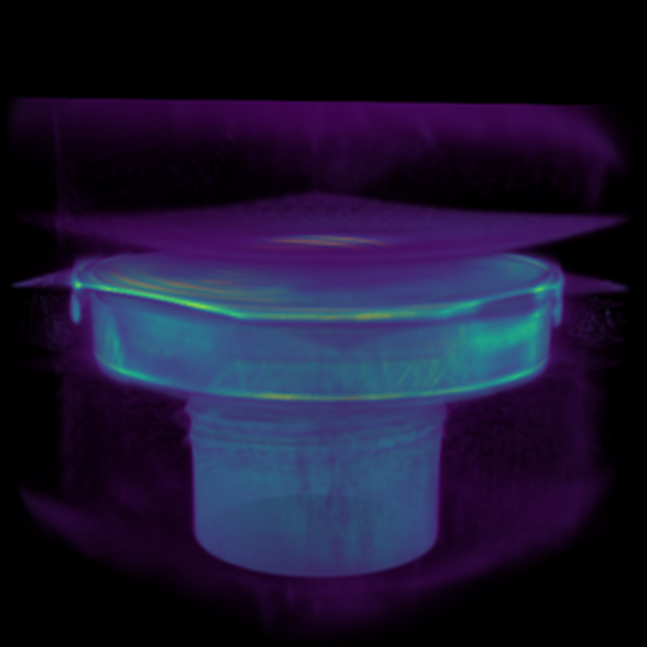}
        & \imgcell{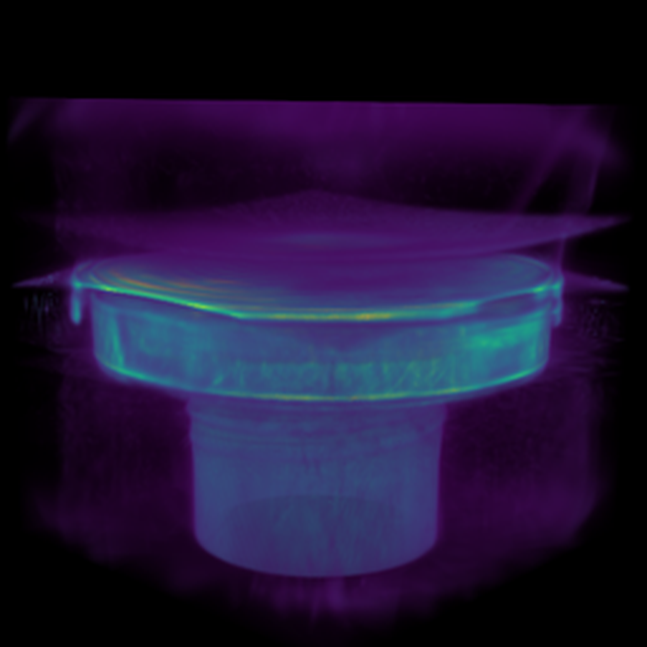}
        \\
        \addlinespace[3pt]

        \rowlabel{120\,keV}{Axial}
        & \imgcell{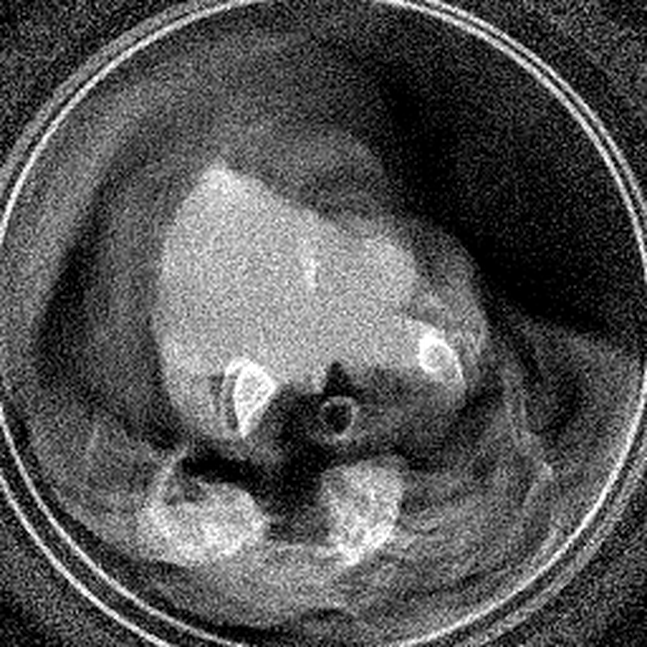}
        & \imgcell{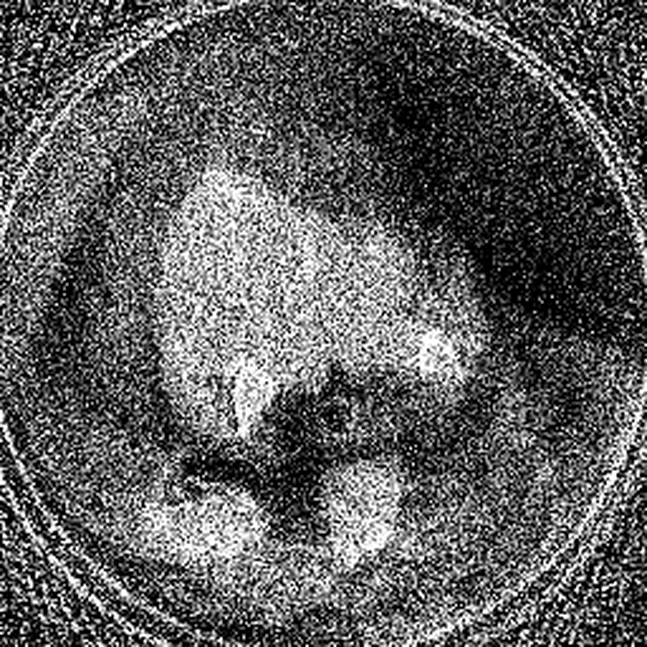}
        & \imgcell{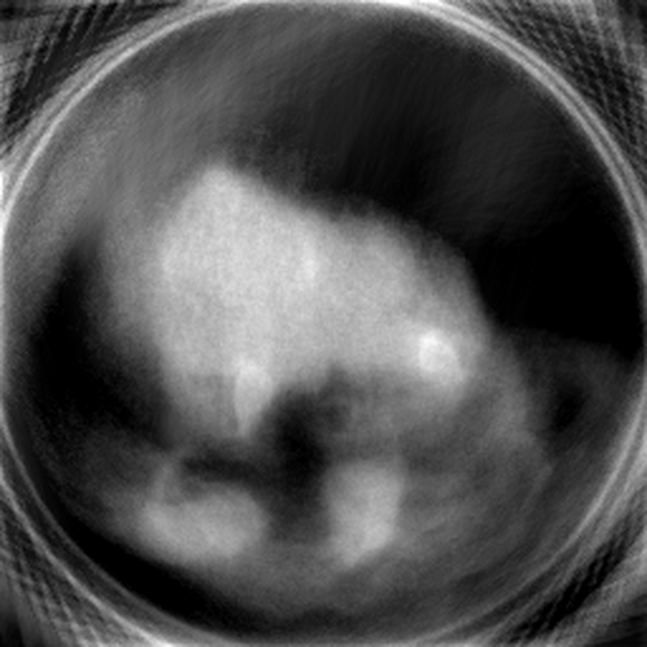}
        & \imgcell{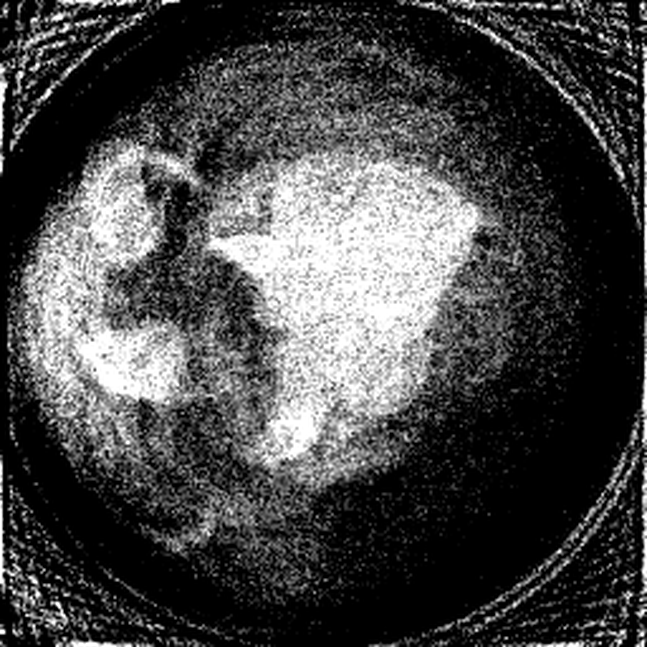}
        & \imgcell{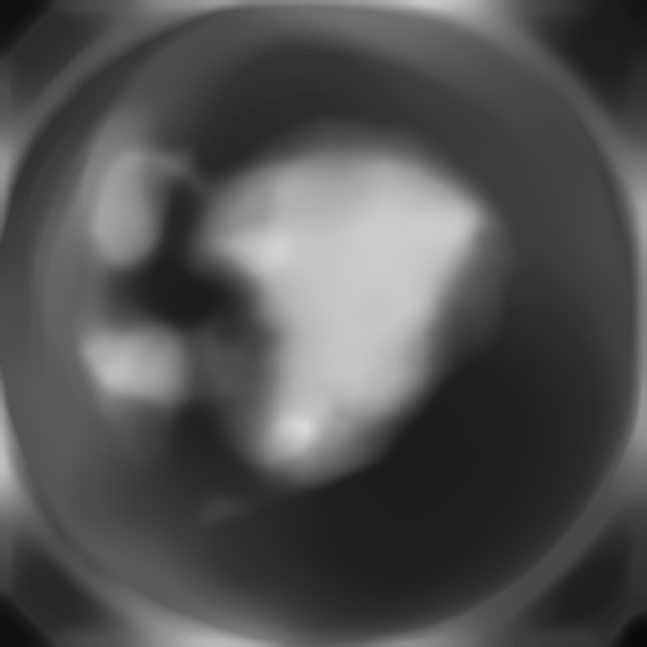}
        & \imgcell{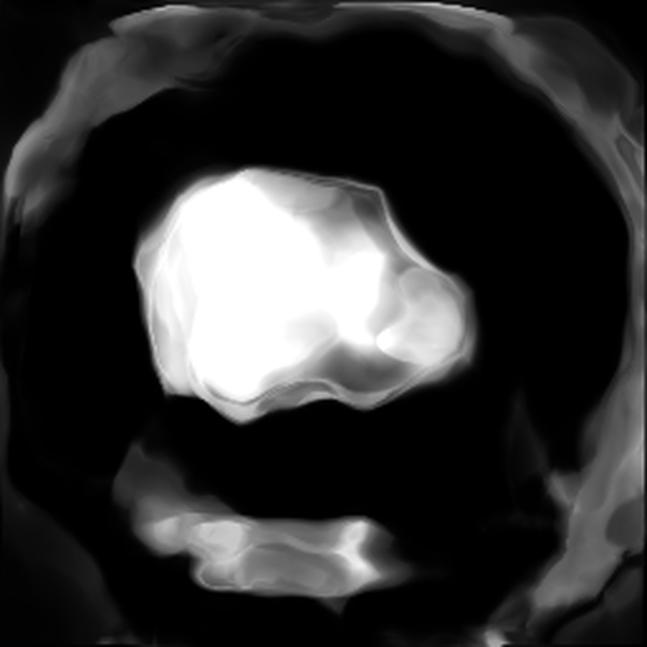}
        & \imgcell{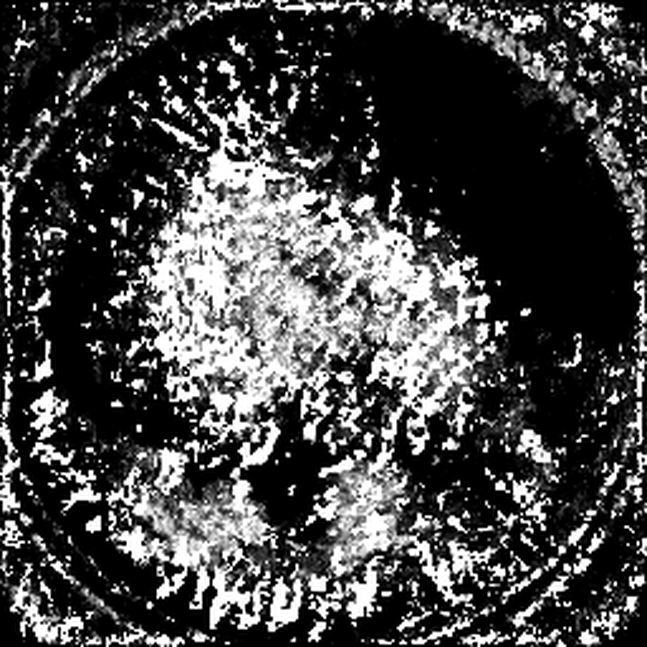}
        & \imgcell{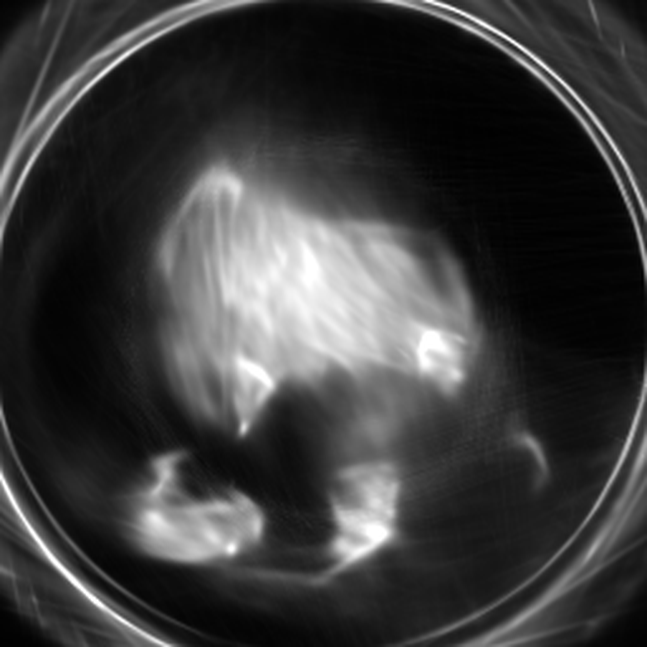}
        & \imgcell{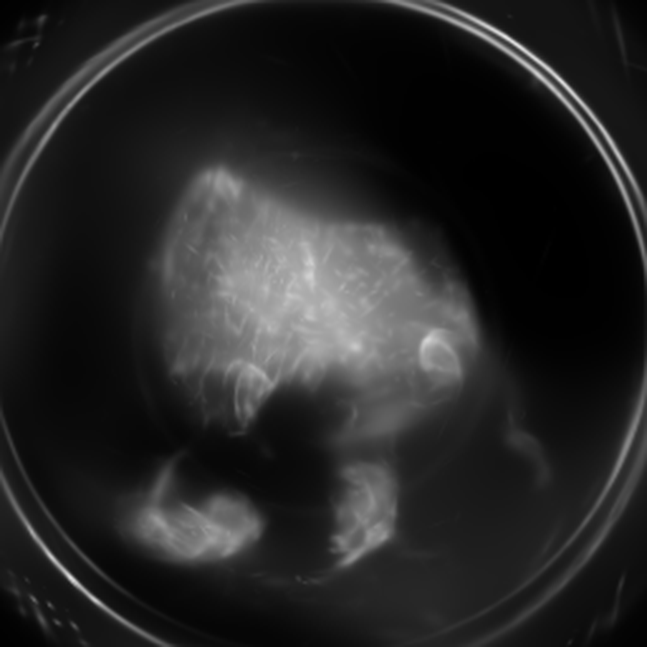}
        & \imgcell{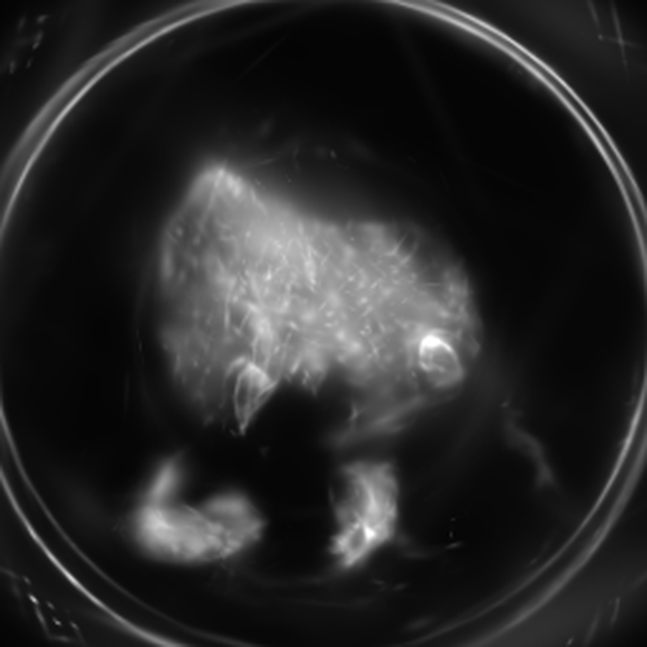}
        \\[-0.5mm]

        \rowlabel{120\,keV}{Outside}
        & \imgcell{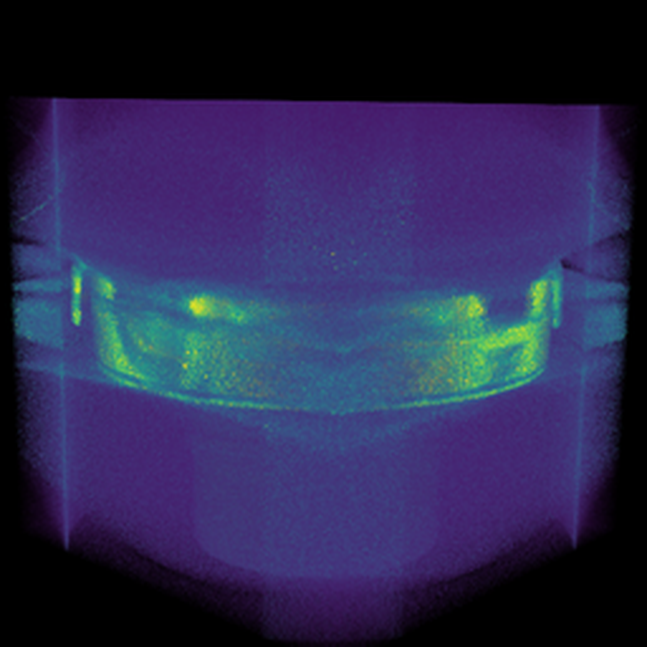}
        & \imgcell{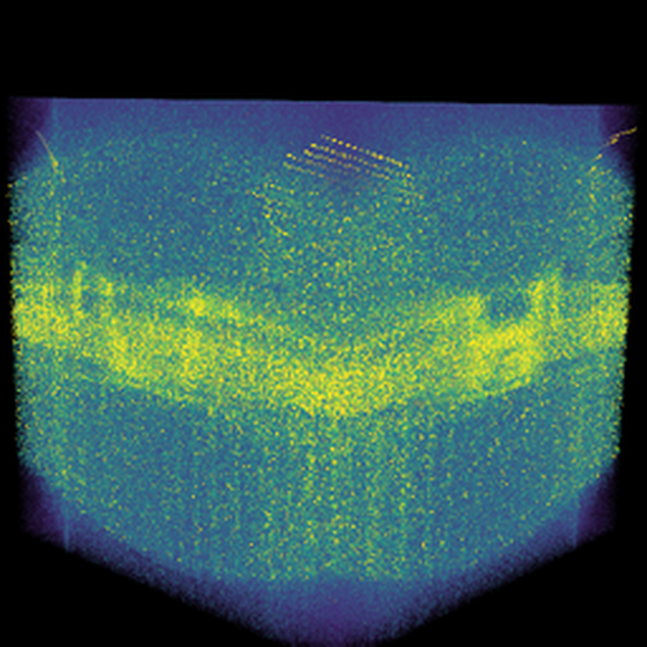}
        & \imgcell{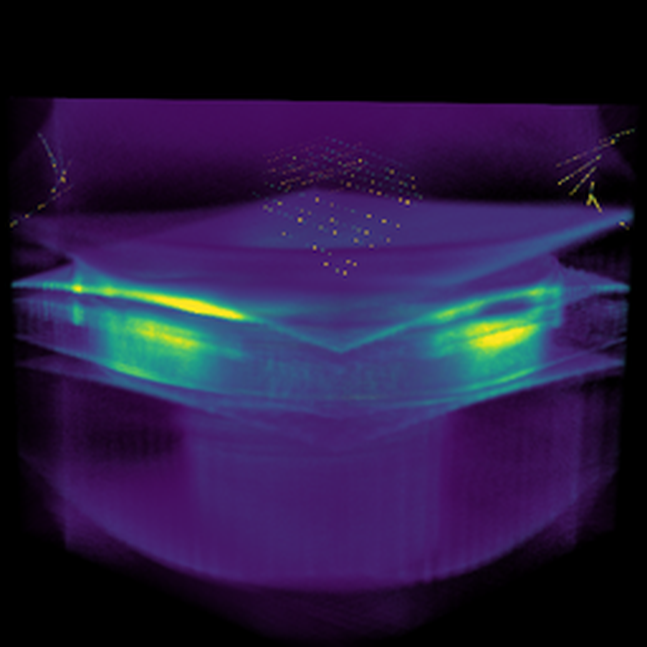}
        & \imgcell{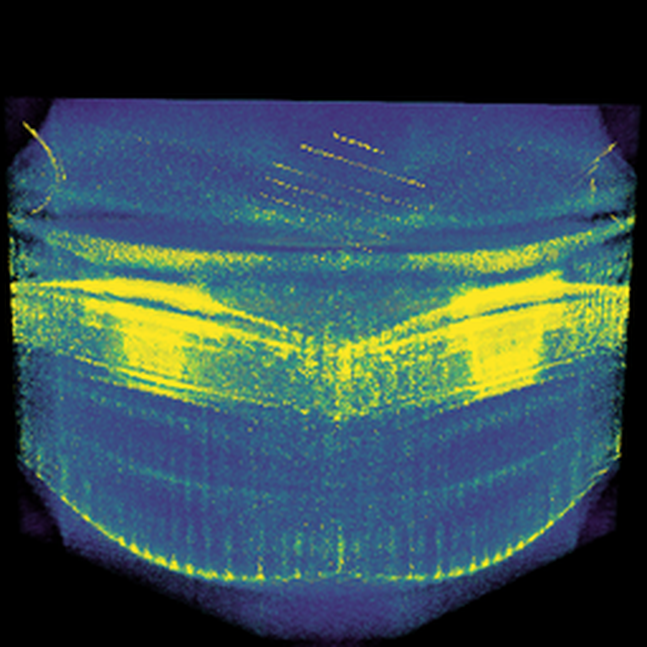}
        & \imgcell{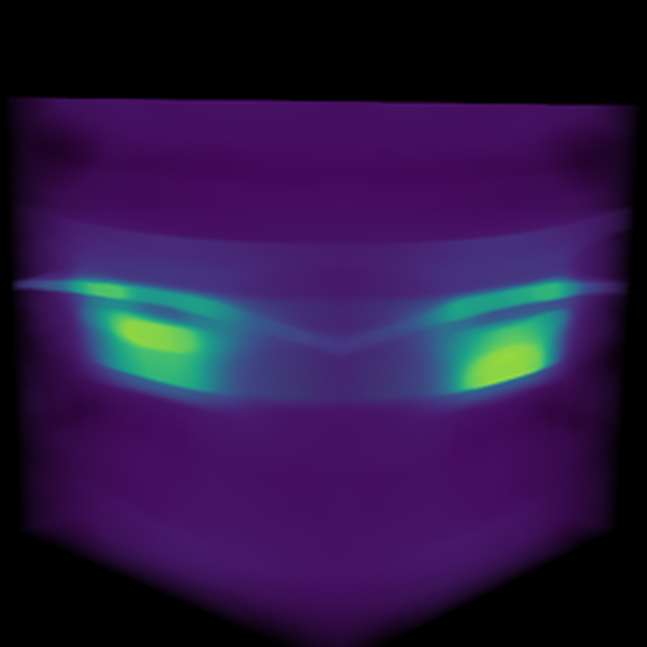}
        & \imgcell{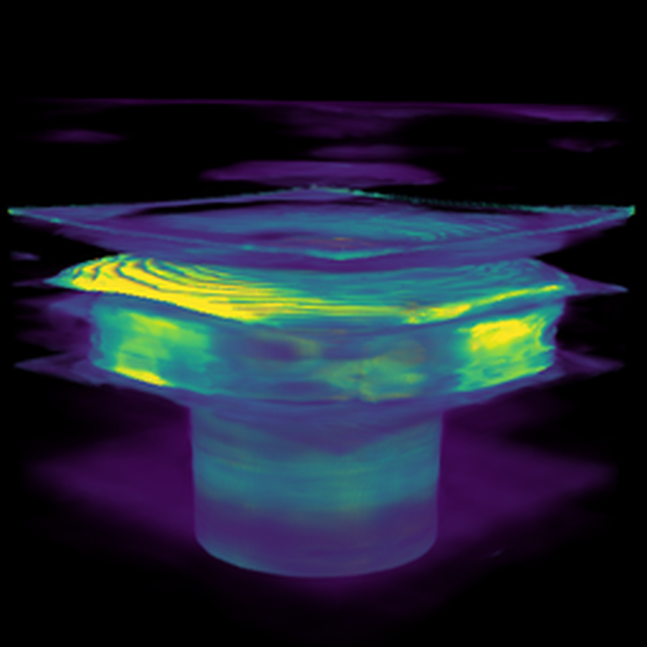}
        & \imgcell{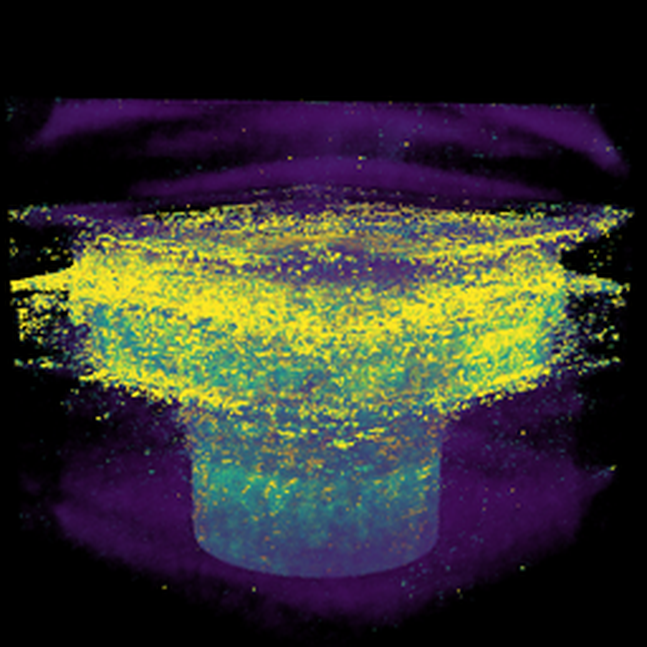}
        & \imgcell{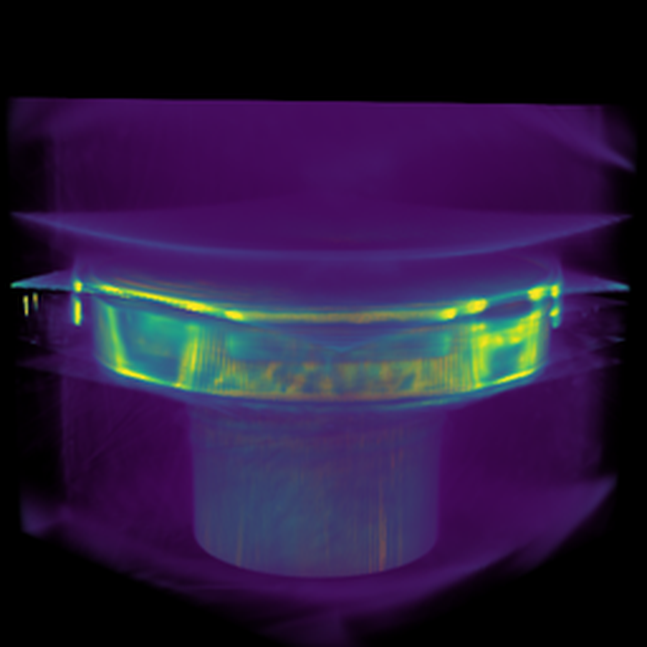}
        & \imgcell{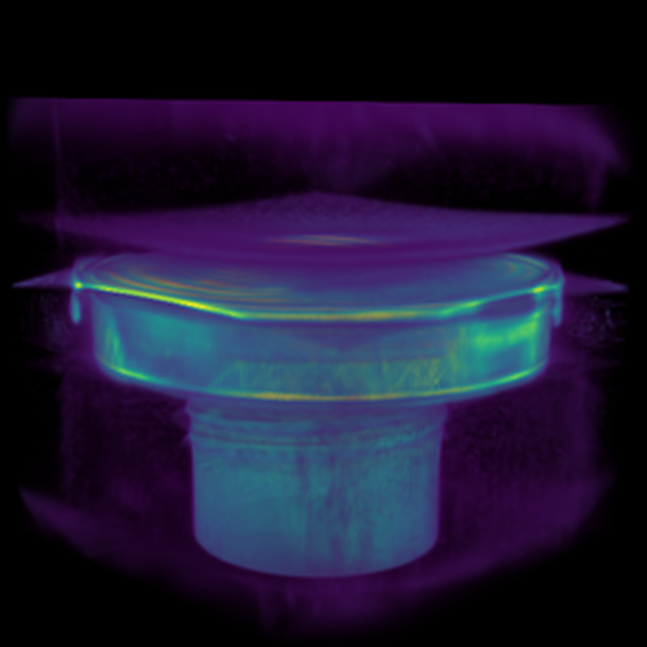}
        & \imgcell{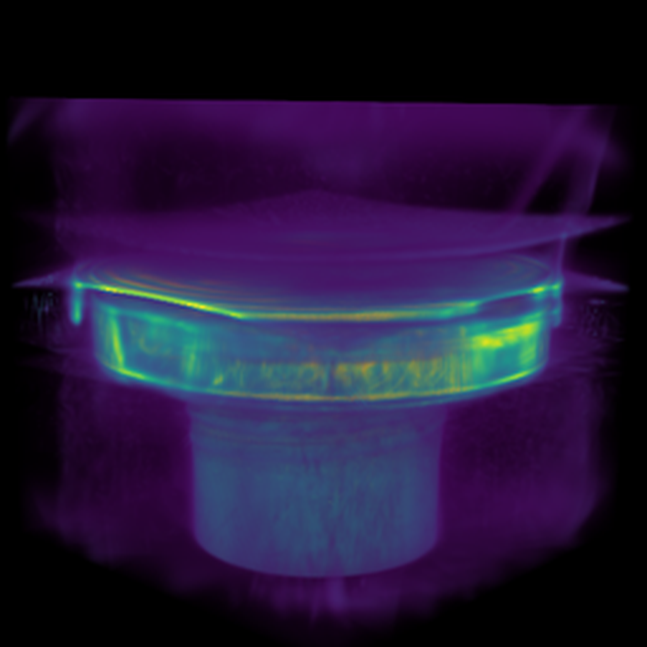}
        \\
        \end{tabular}
\end{adjustbox}}
\end{figure*}
\noindent\textbf{Spectral CT reconstruction}
The lower block of Table~\ref{tab:quail-metrics} evaluates the reconstructed volumes against a
pseudo ground truth, which we obtain by applying FDK~\cite{Feldkamp:84} to all available projections
of each energy channel of the bird chest dataset~\cite{BIRDCHEST_meaney_2024_11388131}, so these numbers should be read as a strong indicator and not as an exact
reference.
In terms of PSNR our method outperforms both traditional solvers and learning-based approaches.
For SSIM and LPIPS we outperform all learning-based methods, with the exception
of R$^2$-Gaussian~\cite{r2_gaussian}, where we reach comparable values while
using only a third of the Gaussians across all energy channels.
Compared to the traditional methods we also reach better SSIM values, and for LPIPS we are better
than all of them except FDK~\cite{Feldkamp:84}, where our values are comparable.
\noindent\textbf{Basis material decomposition}
On the synthetic dataset the photoelectric, Compton and K-edge coefficients are known exactly, so the
recovered basis volumes can be compared directly against the ground truth, and the corresponding
quantitative results are presented in Table~\ref{tab:bmd-synthetic3basis-metrics} and
Table~\ref{tab:bmd-synthetic3basis_rgb-metrics}.
On the photoelectric basis we are better than all other methods in both metrics, with a gain of about
$3$\,dB PSNR and $0.14$ SSIM over the strongest traditional solver, and on the K-edge basis we reach
the best SSIM of all methods (see Table~\ref{tab:bmd-synthetic3basis-metrics}).
We we consider the average over all bases we also reach the best SSIM, so our basis volumes follow the structure of the
ground truth most closely, while the traditional solvers stay ahead in average PSNR, mainly because
of their higher PSNR on the Compton basis.
The material maps, in which every Gaussian is assigned to a material cluster and colored
categorically, cannot be produced by the other methods. We therefore compare our two variants against each other in terms of PSNR, SSIM and LPIPS (see
Table~\ref{tab:bmd-synthetic3basis_rgb-metrics}), where Ours~(3) performs better on all three
metrics, which is expected because the dataset is built from three basis materials.
\subsection{Qualitative results}
\label{sec:qualitative results}
\noindent\textbf{Novel view synthesis}
We now compare the novel view synthesis qualitatively, and the results are presented in
Figure~\ref{fig:nvs-quail-images}, where we show a representative test projection of the bird chest
dataset~\cite{BIRDCHEST_meaney_2024_11388131} for the $50$, $80$ and $120\,\mathrm{keV}$ channels,
together with the absolute error maps with respect to the ground truth.
The images confirm the quantitative findings, since our method and R$^2$-Gaussian~\cite{r2_gaussian} show
the darkest error maps, while X-Gaussian~\cite{cai2024radiativegaussiansplattingefficient} has a bright
error over the whole object and SAX-NeRF~\cite{cai2024structureawaresparseviewxray3d} shows a clear error
in the high-attenuation regions.
For our model this means that the energy spectrum is captured well, because every channel matches the
intensity and the contrast of the ground truth, even though all channels share the same set of Gaussians
and only the energy-dependent basis weights change.
The two variants look almost identical here, which is expected since the object contains no K-edge
material inside the scanned energy range, so the additional basis of Ours~(3) has little to contribute. The remaining error of our method lies at the edges of the bone structures and at the silhouette of the
object, which follows from the Gaussian formulation, where attenuation is represented by smooth density
contributions and sharp interfaces can only be approximated.
The traditional solvers FDK~\cite{Feldkamp:84} and CGLS~\cite{Soleimani2015IterativeCT} instead show a
granular, speckled error over the whole projection, because they reconstruct every channel separately and
therefore also reproduce its noise, whereas our result is visibly smoother, since one shared geometry
cannot fit the independent noise of the channels and thus acts as an implicit denoiser.
\noindent\textbf{Spectral CT reconstruction}
We now qualitatively compare the methods with regards to the reconstructed volumes, for which
Figure~\ref{fig:volume-quail-images} shows an axial slice and an outside view of the full volume at every
energy channel of the bird chest dataset~\cite{BIRDCHEST_meaney_2024_11388131}. It is important to note that since the pseudo ground truth is itself obtained with FDK on all projections, the images are here more
informative than the numbers, which are strongly affected by the absolute density scale and by the noise
in the reference. In the axial slices our reconstruction is clearly closer to the reference than the other learning-based
methods, since it matches the intensity level of the tissue and the bone structures well, while the
remaining methods either stay too dark or brighten towards white and thereby lose the contrast between the
two.
The outside views, however, are difficult for all methods, because the noise in the background is picked up
as spurious density and surrounds the object, so that none of the reconstructions gives a truly clean
surface.
Our approach still produces the most plausible result here, as the shape of the bird chest remains
recognizable and the background stays comparatively uniform, but we emphasize that this is a limitation
that affects our method as well and that none of the compared reconstructions is fully convincing in this
view.
%
%
\begin{figure*}[htbp!]
\centering
\begin{minipage}[t]{0.67\textwidth}
\centering
\vspace{0pt}
\captionof{table}{Ablations on the multi-energy bird chest 
dataset~\cite{BIRDCHEST_meaney_2024_11388131}. \emph{(a)} One energy channel is 
held out from training; the marked channel is the excluded one. 
\emph{(b)} Number of training views for Ours~(3). \emph{(c)} Single components 
of Ours~(3) disabled, relative to the full model.}
\label{tab:ablation-quail-all}
\resizebox{\linewidth}{!}{%
    \setlength{\tabcolsep}{3pt}
    \begin{tabular}{l l ccc ccc ccc}
    \toprule
    \multirow{2}{*}{Ablation} & \multirow{2}{*}{Configuration}
    & \multicolumn{3}{c}{50 keV}
    & \multicolumn{3}{c}{80 keV}
    & \multicolumn{3}{c}{120 keV} \\
    \cmidrule(lr){3-5} \cmidrule(lr){6-8} \cmidrule(lr){9-11}
    & & PSNR$\uparrow$ & SSIM$\uparrow$ & LPIPS$\downarrow$
    & PSNR$\uparrow$ & SSIM$\uparrow$ & LPIPS$\downarrow$
    & PSNR$\uparrow$ & SSIM$\uparrow$ & LPIPS$\downarrow$ \\
    \midrule
    \multicolumn{11}{l}{\textit{(a) Held-out energy channel}} \\
    \midrule
    \multirow{3}{*}{Ours (2)}
    & 50 keV  & \textbf{30.05} & \textbf{0.921} & \textbf{0.291} & 36.67 & 0.946 & 0.260 & 39.48 & 0.947 & 0.311 \\
    & 80 keV  & 33.28 & 0.928 & 0.284 & \textbf{34.79} & \textbf{0.941} & \textbf{0.254} & 39.50 & 0.950 & 0.310 \\
    & 120 keV & 33.46 & 0.929 & 0.281 & 36.21 & 0.949 & 0.254 & \textbf{30.92} & \textbf{0.878} & \textbf{0.321} \\
    \cmidrule(lr){1-11}
    \multirow{3}{*}{Ours (3)}
    & 50 keV  & \textbf{24.99} & \textbf{0.813} & \textbf{0.289} & 37.21 & 0.952 & 0.257 & 39.18 & 0.950 & 0.313 \\
    & 80 keV  & 34.21 & 0.932 & 0.280 & \textbf{30.56} & \textbf{0.924} & \textbf{0.251} & 38.78 & 0.949 & 0.313 \\
    & 120 keV & 34.80 & 0.934 & 0.276 & 36.62 & 0.949 & 0.251 & \textbf{28.22} & \textbf{0.832} & \textbf{0.326} \\
    \midrule
    \multicolumn{11}{l}{\textit{(b) Sparse views}} \\
    \midrule
    \multirow{4}{*}{Ours (3)}
    & 75 (full) & \textbf{33.76} & \textbf{0.930} & 0.282 & \textbf{36.79} & \textbf{0.949} & 0.254 & \textbf{39.03} & \textbf{0.949} & 0.313 \\
    & 50        & 33.63 & 0.929 & 0.281 & 36.61 & 0.947 & 0.253 & 38.86 & 0.948 & \textbf{0.312} \\
    & 25        & 33.00 & 0.922 & \textbf{0.279} & 35.96 & 0.942 & \textbf{0.250} & 38.46 & 0.946 & 0.314 \\
    & 10        & 30.73 & 0.889 & 0.289 & 33.88 & 0.916 & 0.259 & 37.14 & 0.932 & 0.318 \\
    \midrule
    \multicolumn{11}{l}{\textit{(c) Component ablation}} \\
    \midrule
    \multirow{5}{*}{Ours (3)}
    & Ours (3)                            & 33.76 & 0.930 & \textbf{0.282} & 36.79 & 0.949 & \textbf{0.254} & 39.03 & 0.949 & 0.313 \\
    & w/o K-edge freezing             & 33.79 & 0.931 & 0.282 & 36.78 & 0.949 & \textbf{0.254} & 38.95 & 0.949 & 0.314 \\
    & w/o TV loss                     & 33.77 & 0.931 & 0.283 & 36.78 & 0.949 & \textbf{0.254} & 39.01 & \textbf{0.950} & 0.314 \\
    & w/o per-basis densify thresh.  & 33.64 & 0.930 & 0.286 & 36.65 & 0.949 & 0.257 & 38.90 & 0.949 & 0.313 \\
    & w/o polychromatic model     & \textbf{34.31} & \textbf{0.933} & 0.284 & \textbf{37.20} & \textbf{0.951} & 0.257 & \textbf{39.20} & \textbf{0.950} & \textbf{0.312} \\
    \bottomrule
    \end{tabular}}
\end{minipage}\hfill
\begin{minipage}[t]{0.32\textwidth}
\centering
\vspace{20pt} 
\includegraphics[width=\linewidth]{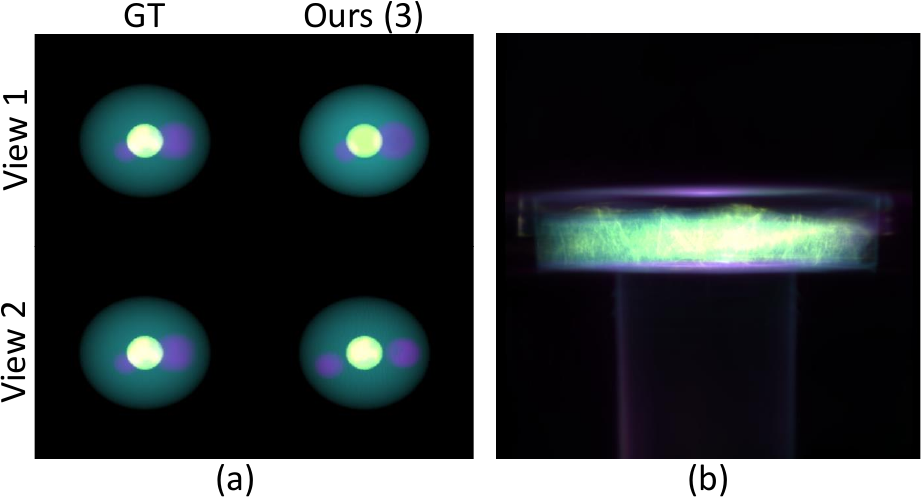}
\caption{Basis material decomposition qualitative results on (a) our synthetic phantom, with ground truth and our prediction for two views, and (b) the real bird chest dataset~\cite{BIRDCHEST_meaney_2024_11388131}, where no ground truth is available. Colors indicate the dominant recovered basis.}
\label{fig:bmd-synthetic3basis_rgb-images}
\end{minipage}
\end{figure*}
\noindent\textbf{Basis material decomposition}
We finally evaluate the basis material decomposition qualitatively on the synthetic phantom dataset that we created for this task.
The predictions are visually very close to the reference (see Figure~\ref{fig:bmd-synthetic3basis_rgb-images}a), since the soft-tissue sphere, the bone cylinder and
both K-edge cylinders appear at the correct location and size and with the correct color, which is consistent
with the reported PSNR and SSIM (see Table~\ref{tab:bmd-synthetic3basis-metrics} and~\ref{tab:bmd-synthetic3basis_rgb-metrics}).
Furthermore, we compare the material decomposition on the real bird chest
dataset~\cite{BIRDCHEST_meaney_2024_11388131} in Figure~\ref{fig:bmd-synthetic3basis_rgb-images}b.
Since no ground-truth material maps are available, this image should be interpreted qualitatively, where the
colors indicate the dominant recovered basis and correct results are expected to show regions that match the
known structures of the dataset, such as separated bone, soft tissue and metal.
The decomposition matches this expectation, since the bone, the soft tissue and the metal structures are
clearly separated and appear at the anatomically plausible locations.  Here, violet indicates metal, greenish tones indicate soft
tissue, and bright regions indicate bone.
The main artifact is an over-saturation of the bone, which becomes overly bright in some regions, so the
material assignment is convincing while the recovered concentration is again less reliable than the spatial
separation.
\subsection{Ablation Studies}
We perform three ablation studies (see Table~\ref{tab:ablation-quail-all}) on the bird chest
dataset~\cite{BIRDCHEST_meaney_2024_11388131}, where we hold out one energy channel from training,
reduce the number of training views, and disable individual components of our method.
In Table~\ref{tab:ablation-quail-all}a, removing the middle channel at $80\,\mathrm{keV}$ costs relatively
little because the model interpolates between two observed energies, while holding out $50$ or
$120\,\mathrm{keV}$ requires extrapolation beyond the trained range and degrades clearly, and Ours~(3) degrades
more than Ours~(2), which indicates that the K-edge basis relies more strongly on channel-specific information.
In Table~\ref{tab:ablation-quail-all}b the quality decreases consistently with fewer training views, but
reducing from $75$ to $25$ views costs less than $1\,\mathrm{dB}$ per channel, and only at $10$ views the
per-view scores spread widely, since test angles close to a training view remain accurate while the remaining
ones drop noticeably.
In Table~\ref{tab:ablation-quail-all}c we start from the full Ours~(3) model in the first row and
disable one of its components in each of the following rows.
Disabling the K-edge freezing, the TV regularization or the per-basis densification thresholds barely
changes the projection metrics, since these components control how the attenuation is split across the
bases while a projection only measures the total attenuation along a ray.
We still keep them, because without them the basis coefficients collapse and the decomposition fails.
The last row replaces the polychromatic energy integral by a single monochromatic evaluation per
channel and scores best, since it can fit each channel independently instead of explaining all
channels with one shared spectrum, which shows that the spectral forward model needs a precise
spectral configuration.

\section{Conclusion}
We presented a spectral CT method that extends 3D Gaussian Splatting with per-Gaussian basis material
fractions and a differentiable polychromatic forward model, so that novel view synthesis, volume
reconstruction and basis material decomposition are obtained from a single optimized representation.
For novel view synthesis, our method outperforms all traditional baselines and stays competitive with per-channel learning-based methods, while using approximately a third of the Gaussians of single-channel approaches.
For volume reconstruction it reaches the best PSNR on the multi-energy bird chest data, and for material
decomposition it is the first method to perform one-step projection-domain decomposition with direct RGB
output, recovering the photoelectric basis more accurately than traditional pipelines.
Our ablations also show that the spectral forward model needs precise calibration and that the decomposition
relies on the stabilizing components, so absolute basis scaling remains the weakest part of the model. Hence, future work could improve the calibration of the forward model, reduce the cross-talk between the
photoelectric and Compton bases, and extend the learnable K-edge component to multiple edges.

\bibliographystyle{unsrt}  
\bibliography{main}  

\end{document}